\documentclass[11pt]{article}

\usepackage[utf8]{inputenc}  
\usepackage[T1]{fontenc}     
\usepackage{amsthm,amsmath,amssymb}
\usepackage{natbib}
\usepackage{multirow}
\usepackage[pdftex]{graphicx}
\usepackage{makecell}
\usepackage{booktabs}
\usepackage{array}
\usepackage{fullpage}
\usepackage{url}
\usepackage{bm}
\usepackage{smile}
\usepackage{mathrsfs}
\usepackage{dsfont}
\usepackage{titling}
\usepackage{epstopdf}
\usepackage{appendix}
\usepackage{enumitem}
\usepackage{microtype}

\usepackage{algorithm}
\usepackage{algpseudocode}
\algtext*{EndFor}
\algtext*{EndIf}
\algtext*{EndWhile}
\algtext*{EndFunction}
\usepackage{pifont}
\usepackage[usenames,dvipsnames,svgnames,table]{xcolor}
\usepackage{tcolorbox}
\tcbuselibrary{breakable}

\usepackage[colorlinks=true,
linkcolor=red,
urlcolor=blue,
citecolor=blue]{hyperref}

\newcommand{\cmark}{{\color{green!70!black}\ding{51}}}
\newcommand{\xmark}{{\color{red!70!black}\ding{55}}}
\newtcolorbox{examplebox}[1]{
  breakable,
  colback=gray!5,
  colframe=gray!60,
  fonttitle=\bfseries\small,
  title=#1,
  boxrule=0.5pt,
  arc=2pt,
  left=6pt, right=6pt, top=4pt, bottom=4pt
}

\makeatletter
\renewcommand\@makefntext[1]{%
  \noindent\hspace*{1em}\@makefnmark\,#1}
\makeatother

\begin{document}

\title{\huge {COOPA: A Modular LLM Agent Architecture for Operations Research Problems}}

\author{
\setcounter{footnote}{0}%
Chuanhao Li$^{1}$\footnotemark[1],
Xiaoan Xu$^{2}$\footnotemark[1],
Dirk Bergemann$^{3}$ \\[0.4em]
Ethan X. Fang$^{2}$, Yehua Wei$^{2}$, Zhuoran Yang$^{3}$ \\[0.6em]
\normalsize $^{1}$Tsinghua University \quad $^{2}$Duke University \quad $^{3}$Yale University
}
\date{}
\maketitle
\renewcommand{\thefootnote}{\fnsymbol{footnote}}%
\footnotetext[1]{These authors contributed equally to this work.}%
\renewcommand{\thefootnote}{\arabic{footnote}}%

\begin{abstract}
Operations Research (OR) provides a rigorous framework for high-stakes decision-making, but effective OR modeling requires substantial domain knowledge, mathematical abstraction, and solver expertise. Recent LLM-based systems automate parts of this pipeline, yet remain limited by low accuracy on complex problems, opaque outputs, and narrow solver support. We propose COOPA (\textbf{CO}operative \textbf{OP}erations \textbf{A}gent), a modular LLM-agent architecture for interpretable and scalable OR decision support. It combines three components: iterative confidence-based modeling, which generates multiple candidate formulations, self-evaluates them across modeling dimensions, and selects one using a max-min confidence criterion; element-level provenance and confidence explanations, which link variables, parameters, constraints, and objectives to quoted source text and provide an audit trail for human verification; and multi-solver routing to specialized optimizer agents for different OR problem classes. Across three OR benchmarks, eight LLM backbones, and four baselines under identical conditions, COOPA achieves the best macro-average accuracy on six of eight backbones and improves over the strongest baseline by up to 6.7 percentage points. A within-system ablation isolates the contribution of iterative confidence-based modeling, while additional analyses and case studies illustrate the value of source traceability and multi-solver dispatch.\footnote{All resources are available at \url{https://github.com/xxxxxa-hub/COOPA}.}
\end{abstract}


\section{Introduction}



Operations Research (OR) provides a rigorous foundation for decision-making in complex systems. Across domains such as supply chains, transportation, energy, healthcare, finance, engineering, and public policy, OR uses mathematical and computational models to capture resource constraints, system dynamics, uncertainty, and performance objectives. Its toolkit spans deterministic, stochastic, and robust optimization; dynamic programming; queueing theory; and simulation-based optimization, supported by exact algorithms, decomposition methods, approximation algorithms, heuristics, and metaheuristics. This breadth has made OR a foundational discipline for high-stakes decision support.


However, applying OR effectively remains expertise-intensive. Building a useful model requires specifying variables, objectives, and constraints; deciding what to model explicitly or abstract away; representing uncertainty, temporal structure, and system interactions; balancing decision relevance with computational tractability; and implementing the model in specialized optimization environments. This process is typically iterative and requires close interaction with domain experts. Even experienced practitioners may spend weeks or months refining formulations, diagnosing computational behavior, and validating results against operational reality. Consequently, OR remains underutilized in many organizations, especially where specialized modeling expertise is scarce.


Recent advances in large language models (LLMs) offer a plausible way to lower this barrier. LLMs can process natural-language problem descriptions, generate structured outputs and code, and reason over symbolic representations, making them natural candidates for tasks such as formulation, solver-oriented code generation, and result interpretation. A growing literature explores this direction through prompting, code synthesis, and fine-tuning strategies tailored to optimization tasks \cite{xiao2023chain,ahmaditeshnizi2024optimus,tang2024orlm,jiang2024llmopt}. Yet current systems remain far from reliable OR decision-support tools. We highlight three limitations.


First, \textit{accuracy on harder OR problems remains limited, and formulation errors are the main bottleneck}. Across several studies, incomplete formulations, incorrect variables or constraints, and misspecified objectives account for 50--70\% of failures, substantially more than coding errors \cite{tang2024orlm, liu2025optitree, ahmaditeshnizi2024optimus}. Most methods still generate a single formulation in one pass, with little opportunity to improve it before code generation \cite{xiao2023chain, liu2025optitree, zhang2025or, tang2024orlm}. When revision occurs, it is usually triggered by downstream execution failures rather than explicit iteration on the formulation itself \cite{ahmaditeshnizi2024optimus, jiang2024llmopt}. Thus, mathematically incorrect but executable formulations can survive unchecked.

Second, \textit{existing systems produce opaque outputs}. Most LLM-based OR systems present users with a formulation, solver code, and an optimal solution value \cite{ahmaditeshnizi2024optimus, xiao2023chain, zhang2025or, tang2024orlm}; some additionally report quality indicators such as per-clause confidence scores \cite{ahmaditeshnizi2024optimus}. However, none of these outputs includes the rationale behind quality assessments, such as why a particular element received a given score or what aspect of the problem description it was derived from, and no system provides element-level provenance linking each parameter, variable, constraint, or objective term back to the specific text that motivated it. In practical OR problems involving hundreds of constraints, the absence of such traceability forces practitioners to re-read the entire problem description and mentally reconstruct the LLM's reasoning to verify correctness or locate errors, which is labor-intensive and does not scale.


Third, \textit{existing systems are difficult to adapt to new solver backends}. For methods that rely on model fine-tuning, solver dependence is direct: ORLM \cite{tang2024orlm}, OR-R1 \cite{ding2025or}, and SIRL \cite{chen2025solver} target COPT, while LLMOPT \cite{jiang2024llmopt} uses Pyomo. Changing the backend in such methods requires new solver-specific data, revised output formats, or additional fine-tuning. Systems without fine-tuning avoid solver-specific retraining, but solver-specific assumptions can still be embedded throughout the workflow, including formulation templates, code-generation prompts, solver API calls, execution and debugging logic, and output parsing. OptiMUS \cite{ahmaditeshnizi2024optimus}, CoE \cite{xiao2023chain}, OptiTree \cite{liu2025optitree}, OR-LLM-Agent \cite{zhang2025or}, and LEAN-LLM-OPT \cite{liang2026llm}, for example, generate Gurobi-oriented code. OptimAI \cite{thind2025optimai} broadens coverage by supporting multiple mathematical-programming backends, including PuLP, Pyomo, and OR-Tools. Nevertheless, existing systems generally treat backend support as a fixed implementation choice rather than an extensible architectural interface. 

Motivated by these limitations, we propose COOPA (\textbf{CO}operative \textbf{OP}erations \textbf{A}gent), a modular LLM-agent architecture for OR modeling and solver execution built around three design requirements: improving formulation quality before code generation, representing modeling artifacts in a structured and traceable form, and supporting extensible solver dispatch. COOPA implements these requirements through three components. First, \emph{iterative confidence-based modeling}: it generates multiple candidate formulations, scores each across four modeling dimensions with confidence explanations, and selects the candidate with the highest minimum confidence via a max-min criterion. Second, \emph{interpretable output for human verification}: it extracts modeling elements through a validated schema and links them to quoted source text, creating an audit trail from problem statement to formulation. Third, \emph{scalable multi-solver dispatch}: it classifies the problem type and routes it to one of four optimizer agents covering Pyomo, OR-Tools, pymoo, and standard Python; new agents can be added without modifying the rest of the workflow.

We evaluate COOPA on three benchmarks and eight LLM backbones against four agentic baselines. COOPA achieves the highest macro-average accuracy on 6 of 8 backbones. The strongest results are COOPA with GPT-5.2 (70.6\%), GPT-5 (69.4\%), and Gemini-3-Flash (68.4\%), suggesting that the architecture amplifies strong backbones. The gains also extend beyond the very top models: COOPA improves over the best baseline by 6.7 percentage points on GPT-5 and by 6.3 on GPT-4.1. A within-system ablation then isolates the contribution of iterative modeling.


\section{Failure Modes and Design Requirements for LLM-Based OR Systems} \label{sec:design_choices}


We provide a detailed discussion of related work in Section~\ref{sec:related_work} and focus here on the failure modes that motivate COOPA. The key bottleneck of LLM-based OR systems is often the \emph{problem-to-model} stage: translating natural language into variables, objectives, constraints, and parameters. This motivates two requirements for COOPA: improving formulation quality before code generation and representing modeling artifacts in a traceable form. A third requirement, extensible solver dispatch, follows from the broader diversity of OR problem classes and solver backends.

\subsection{Formulation Errors as the Central Bottleneck}
\label{sec:modeling_errors}


Existing analyses consistently identify formulation-stage errors as a major source of failure. ORLM's error taxonomy \cite{tang2024orlm} concludes that ``the primary challenges lie in the optimization modeling phase,'' while code generation achieves a pass rate of approximately 95\%. Among sampled modeling errors, 56.3\% come from low model completeness, 30.3\% from objective or constraint translation errors, and 13.4\% from semantic misunderstanding of the problem text. OptiMUS \cite{ahmaditeshnizi2024optimus} similarly reports that, on hard mixed-integer problems with multi-dimensional variables, parameter extraction and modeling are harder than coding the resulting model. OptiTree \cite{liu2025optitree} further observes that many medium- and hard-instance failures stem from incorrect variable definitions.
These findings imply that code-generation improvements alone are insufficient. Execution feedback can catch syntax errors, runtime exceptions, or malformed constraints, but not formulations that are executable yet semantically inconsistent with the problem description. COOPA therefore targets formulation quality, using iterative candidate generation, confidence-based assessment, and structured extraction.

\subsection{From Failure Modes to Design Requirements}
\label{sec:diverse_approaches}

LLM-based OR systems differ both in how they construct formulations and in which solver backends they support. These workflow choices point to three design requirements for COOPA.

\textbf{Pre-code formulation improvement.} Methods differ both in how they obtain the formulation passed to code generation and in when they verify it. Some generate an initial formulation and rely on downstream feedback to repair errors, using limited retries \cite{xiao2023chain, zhang2025or} or correction loops driven by execution or solver feedback \cite{ahmaditeshnizi2024optimus, jiang2024llmopt}. Others improve formulation quality through taxonomy-guided decomposition \cite{liu2025optitree} or fine-tuning on OR tasks \cite{tang2024orlm, chen2025solver, ding2025or}. These approaches can help, but they usually revise a single formulation trajectory rather than explicitly comparing alternatives before code generation. Moreover, verification is often downstream- or outcome-driven: some systems use evaluate-improve cycles \cite{ahmaditeshnizi2024optimus}, execution-based self-correction \cite{jiang2024llmopt}, or backward reflection after solution failure \cite{xiao2023chain}, while others accept the first result once the generated code runs and returns an answer \cite{zhang2025or, liu2025optitree, tang2024orlm}. This leaves a key gap: a formulation can be executable and still misrepresent the original problem semantics. COOPA therefore evaluates formulation quality before solver code is generated, generating multiple candidates and selecting among them with confidence-based assessment.

\textbf{Structured and traceable modeling artifacts.} Systems also differ in how they represent intermediate artifacts. Prior work uses JSON \cite{ahmaditeshnizi2024optimus}, regex extraction from solver logs \cite{liu2025optitree}, free-form comments \cite{xiao2023chain}, and Markdown outputs \cite{zhang2025or}. This choice matters because downstream stages must reliably recover variables, parameters, constraints, and objectives from LLM outputs. Brittle formatting assumptions or ad hoc parsing can therefore affect both execution and measured accuracy; for example, we find that the regex-based extraction used by OptiTree \cite{liu2025optitree} substantially reduces accuracy on some backbones relative to LLM-based extraction. At the same time, existing systems provide limited support for human verification: they return a formulation, solver code, and final answer, but not element-level provenance linking variables, parameters, constraints, or objective terms to the problem text, nor confidence information for individual modeling choices. Users must therefore reconstruct why each modeling element was introduced, a process that becomes costly as formulations grow larger. COOPA addresses both issues with schema-validated structured outputs and source traceability.


\textbf{Extensible solver dispatch.} Solver support matters not only in the range of backends currently available, but also in how easily the architecture can be extended to new ones. OR problems may require mathematical programming, constraint programming, multi-objective optimization, simulation-based modeling, or black-box search. Systems with broader backend coverage, such as OptimAI \cite{thind2025optimai}, still focus mainly on mathematical-programming interfaces. COOPA instead separates problem classification from solver-specific optimization: it routes each instance to a specialized optimizer agent, and new optimizer agents can be added without changing the rest of the workflow.




The three requirements above motivate the three components in Section~\ref{sec:method}: structured modeling with source traceability, iterative confidence-based model selection, and multi-solver dispatch.

\begin{figure}[t]
    \centering
    \includegraphics[width=0.9\textwidth]{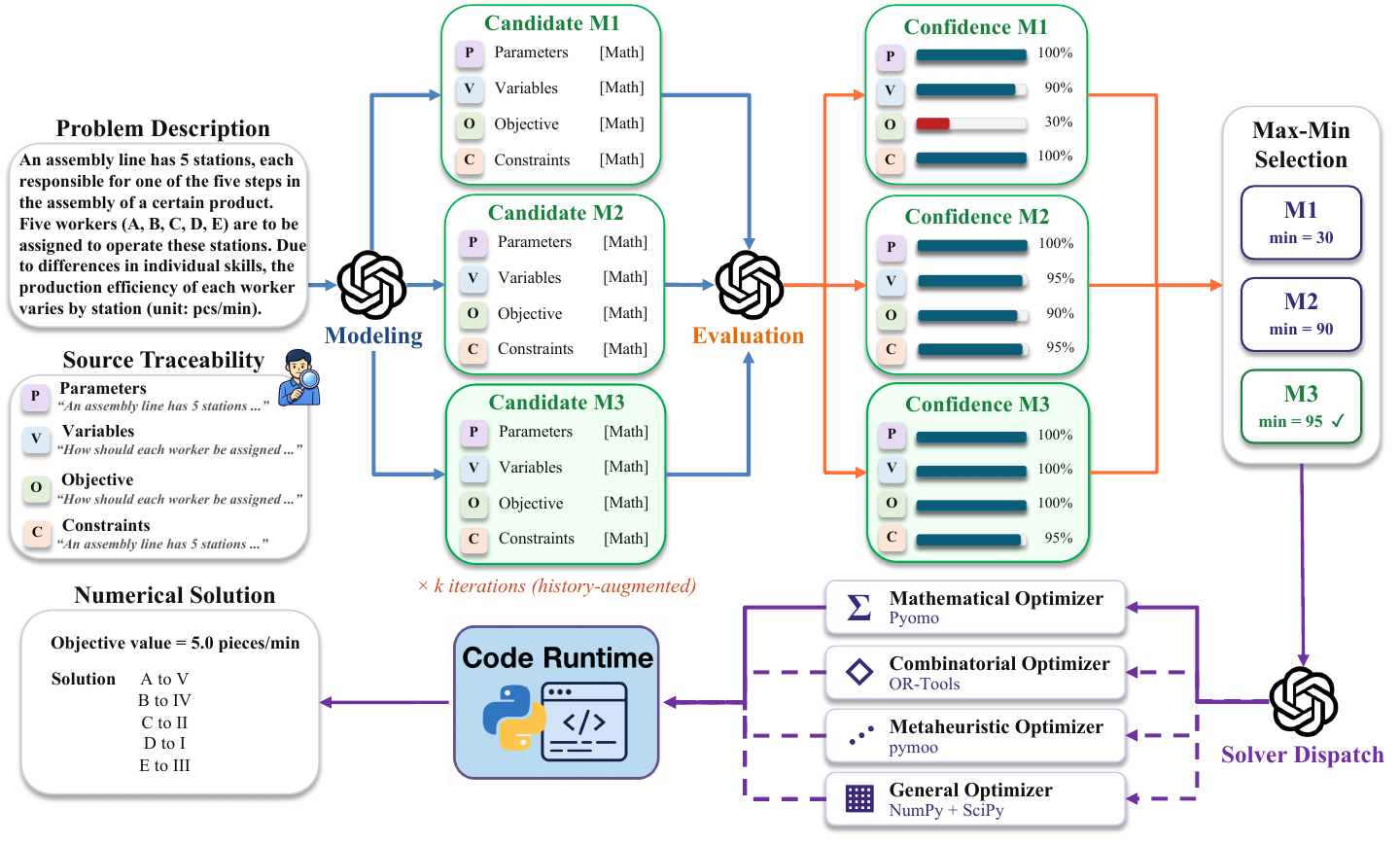}
    \caption{Overview of COOPA. Problem description is parsed into structured candidate models with source traceability, filtered by confidence-based selection, and dispatched to a specialized solver.}
    \label{fig:architecture}
\end{figure}

\section{The Architecture of COOPA} \label{sec:method}

COOPA addresses the three design requirements through structured modeling with source traceability (Section~\ref{sec:structured_modeling}), iterative confidence-based selection (Section~\ref{sec:iterative_modeling}), and solver dispatch (Section~\ref{sec:solver_dispatch}). 

\subsection{System Overview} \label{sec:system_overview}

COOPA is a hierarchical multi-agent system implemented with \texttt{smolagents} \cite{smolagents}. It is organized around three components, shown in Figure~\ref{fig:architecture} and listed below. Prompts are given in Appendix~\ref{sec:appendix_prompts}.

\begin{enumerate}[leftmargin=*,noitemsep]
    \item \textbf{Structured modeling with source traceability.} The problem is converted into a structured candidate model with parameters, decision variables, an objective, and constraints. Each element is produced under a Pydantic schema and includes a \emph{source field} quoting the original problem text.
    \item \textbf{Iterative confidence-based model selection.} The structured-modeling step is repeated $k$ times, with $k=3$ by default, to obtain multiple candidate models. For each candidate, the LLM assigns confidence scores from 0 to 100 across four modeling dimensions and provides short explanations. The candidate that maximizes the minimum confidence score across dimensions is selected.
    \item \textbf{Multi-solver dispatch.} The selected model is then routed to one of the optimizer agents. Each optimizer agent is an LLM controller specialized to a solver package: it translates the model into solver-specific code, invokes the corresponding backend, and returns the executed result.
\end{enumerate}
We use the following BWOR Question~76 \cite{zhang2025or} as a running example to illustrate COOPA. The reported candidate models, confidence scores, and solver outputs come from an actual run (o4-mini backbone). The key challenge here is that assembly-line throughput is determined by the \emph{slowest} station, so the formulation must optimize a bottleneck objective rather than total station output.

\begin{examplebox}{Running Example: Problem Description (BWOR Question 76)}
\small
\begin{minipage}[t]{0.6\textwidth}
An assembly line has 5 stations, each responsible for one of the five steps in the assembly of a certain product. Five workers (A, B, C, D, E) are to be assigned to operate these stations. Due to differences in individual skills, the production efficiency of each worker varies by station (unit: pieces/min). 

\textbf{Question:} How should each worker be assigned to a station to maximize the overall production capacity of the assembly line?

\textbf{Gold answer:} $T^* = 5.0$ pieces/min (bottleneck throughput).
\end{minipage}
\hfill
\begin{minipage}[t]{0.36\textwidth}
\centering
\vspace{-0.4em}
\begin{tabular}{lccccc}
\toprule
\textbf{Worker} & \textbf{I} & \textbf{II} & \textbf{III} & \textbf{IV} & \textbf{V} \\
\midrule
A & 2 & 3 & 4 & 1 & 7 \\
B & 3 & 4 & 2 & 5 & 6 \\
C & 2 & 5 & 3 & 4 & 1 \\
D & 5 & 2 & 3 & 2 & 5 \\
E & 3 & 7 & 6 & 2 & 4 \\
\bottomrule
\end{tabular}
\end{minipage}


\end{examplebox}

\subsection{Structured Modeling with Source Traceability} \label{sec:structured_modeling}

The first component, shown on the left of Figure~\ref{fig:architecture}, defines the structured representation used throughout the pipeline and applies it once to the problem description to produce a candidate model whose elements can be checked individually. A predefined Pydantic schema ensures that downstream components receive structured artifacts in a consistent form.
Each model contains four element types:
\begin{itemize}[leftmargin=*,noitemsep]
    \item \texttt{ParameterDefinition}: name, symbol, value or domain, and source reference.
    \item \texttt{VariableDefinition}: name, symbol, type, bounds, and source reference.
    \item \texttt{ObjectiveDefinition}: direction, math expression, variables involved, and source reference.
    \item \texttt{ConstraintDefinition}: name, math expression, type, and source reference.
\end{itemize}
Each element includes a required source reference quoting the original problem text that supports it. This makes the formulation easier to inspect and helps localize errors when the final answer is wrong. Schema validation reduces reliance on post-hoc parsing and ensures that downstream stages receive well-formed objects with the required fields. Validation guarantees structure, not mathematical correctness; correctness is assessed separately in Section~\ref{sec:iterative_modeling}.

The following box shows selected elements from the first candidate produced with this schema in the running example. Each element includes a source reference linking it to the original problem text.

\begin{examplebox}{Running Example: Candidate Model $M_{1}$}
\small
\setlength{\tabcolsep}{4pt}
\renewcommand{\arraystretch}{1.12}

\begin{center}
\begin{tabular}{@{}p{0.16\textwidth}p{0.27\textwidth}p{0.47\textwidth}@{}}
\toprule
\textbf{Type} & \textbf{Element} & \textbf{Source reference} \\
\midrule
Parameters
& \texttt{num\_stations}=5
& ``An assembly line has 5 stations\ldots'' \\
& \texttt{num\_workers}=5
& ``Five workers --- A, B, C, D, and E --- are to be assigned to operate these stations.'' \\
& \texttt{efficiency\_A\_I}=2
& ``The specific efficiencies are shown in Table 8--10 \ldots''; worker A, station I \\
& \texttt{efficiency\_B\_IV}=5
& ``The specific efficiencies are shown in Table 8--10 \ldots''; worker B, station IV \\
& $\cdots$
& $\cdots$ (23 more efficiency parameters) \\
\addlinespace[3pt]

Variables
& $x_{w,s}\in\{0,1\}\quad \forall w,s$
& ``How should each worker be assigned to a station to maximize the overall production capacity of the assembly line?'' \\
\addlinespace[3pt]

Objective
& $\max \sum_{w,s} e_{w,s}x_{w,s}$\quad \colorbox{red!15}{\textbf{WRONG}}
& ``How should each worker be assigned to a station to maximize the overall production capacity of the assembly line?'' \\
\addlinespace[3pt]

Constraints
& $\sum_s x_{w,s}=1\quad \forall w$
& ``Five workers --- A, B, C, D, and E --- are to be assigned to operate these stations.'' \\
& $\sum_w x_{w,s}=1\quad \forall s$
& ``An assembly line has 5 stations\ldots'' \\
\bottomrule
\end{tabular}
\end{center}
\end{examplebox}


This highlights both the value and the limit of source traceability. The binary variables that represent assigning workers to stations, as well as the constraints that assign each worker to one station and each station to one worker, are well grounded in the quoted text. The objective, however, cites the phrase ``maximize the overall production capacity'' but incorrectly encodes it as maximizing the sum of station efficiencies rather than the bottleneck throughput. Source references therefore make modeling choices easier to audit, but do not by themselves ensure correctness. The next component builds directly on this one: it generates several candidate models in the same schema and then evaluates them against the original problem statement.

\subsection{Iterative Confidence-Based Model Selection} \label{sec:iterative_modeling}

Because LLMs sometimes misinterpret parts of the problem description, committing to a single candidate leaves no opportunity to catch errors before code generation. As shown in the center of Figure~\ref{fig:architecture}, COOPA therefore invokes the structured-modeling component multiple times and uses self-evaluated confidence to select the most reliable candidate before dispatch.

\begin{algorithm}[t]
\caption{Iterative Confidence-Based Model Selection}
\label{alg:iterative}
\begin{algorithmic}[1]
\Statex \textbf{Input:} Problem description $P$, number of candidates $k$
\State $\mathcal{H}\gets\emptyset$ 
\Comment{History of candidates and evaluations}
\For{$i=1,\ldots,k$}
    \State $M_i\gets\textsc{GenerateModel}(P,\mathcal{H})$ \Comment{Structured model with source references}
    \State $(\{c_{i,j}\}_j,\{e_{i,j}\}_j)\gets\textsc{EvaluateConfidence}(P,M_i)$ \Comment{One call returns all scores}
    \State $\mathcal{H}\gets\mathcal{H}\cup\{(M_i,\{c_{i,j},e_{i,j}\}_{j})\}$
\EndFor
\State $M^*\gets\arg\max_{i\in\{1,\ldots,k\}}\min_j c_{i,j}$ \Comment{Max-min selection}
\State \Return $M^*$
\end{algorithmic}
\end{algorithm}

The procedure operates as follows (Algorithm~\ref{alg:iterative}). In this component, the LLM first invokes the structured-modeling component to produce a candidate model $M_1$ under the same schema, with source references attached to each element, then assigns confidence scores $c_{1,j} \in [0,100]$ across four dimensions: parameters, variables, objective, and constraints, together with short explanations. The evaluation uses one structured output call over the candidate and the original problem text; no external rubric or few-shot examples are provided. The LLM then generates revised candidates $M_2, \ldots, M_k$ using the original problem together with earlier candidates and their evaluations, allowing low-confidence dimensions to be addressed explicitly.
After $k$ candidates have been generated and evaluated (default $k=3$), the system selects the final model using a \emph{max-min} strategy: $M^* = \arg\max_{i \in \{1, \ldots, k\}} \min_{j \in \{1,2,3,4\}} c_{i,j}$.
That is, the selected model is the one with the highest worst-case confidence across the four modeling dimensions.
This targets the common case where one flawed element invalidates an otherwise strong formulation, consistent with the failure mode discussed in Section~\ref{sec:modeling_errors}. These scores also support human verification by indicating uncertain dimensions; we discuss their calibration in Appendix~\ref{sec:disc_confidence}.

The next two boxes illustrate how the confidence scores guide model selection. The first shows the evaluation of the initial candidate $M_{1}$. The second shows the formulation selected as $M^{\star}$.

\begin{examplebox}{Running Example: Confidence Evaluation of Candidate Model $M_{1}$}
\small
\setlength{\tabcolsep}{4pt}
\renewcommand{\arraystretch}{1.12}

\noindent\makebox[\textwidth][c]{%
\begin{tabular}{@{}p{0.1\textwidth}p{0.1\textwidth}p{0.77\textwidth}@{}}
\toprule
\textbf{Type} & \textbf{Score} & \textbf{Comments} \\
\midrule
Parameters
& 95/100
& All station and worker counts and the 25 efficiency parameters are identified with correct values and units. \\
Variables
& 100/100
& Binary variable x[w,s] is defined for each worker--station pair with correct domain. \\
Objective
& \makecell[tl]{30/100\\[-2pt]\colorbox{red!15}{\scriptsize\textbf{MIN}}}
& The formulation maximizes the sum of efficiencies, but assembly-line capacity is driven by the bottleneck (minimum station rate), so the objective is mis-specified. \\
Constraints
& 100/100
& Exactly-one assignment constraints for each worker and each station are correctly specified. \\
\bottomrule
\end{tabular}
}
\end{examplebox}

\noindent The scores isolate the objective as the only weak dimension: it receives 30/100, while parameters, variables, and constraints score 95--100. This indicates that the main error lies in how throughput is formulated, not in the extracted entities or assignment structure. Guided by the low objective score, $M_{2}$ replaces the incorrect sum-of-efficiencies objective with a bottleneck formulation, and $M_{3}$ refines it further. The selected model $M^{\star}=M_{3}$ makes three key changes relative to $M_{1}$ as shown below.

\begin{examplebox}{{Running Example: Changes of Selected Candidate Model $M^{\star}$ Relative to $M_{1}$}}
\small

\setlength{\tabcolsep}{4pt}
\renewcommand{\arraystretch}{1.12}

\noindent\makebox[\textwidth][c]{%
\begin{tabular}{@{}p{0.16\textwidth}p{0.30\textwidth}p{0.44\textwidth}@{}}
\toprule
\textbf{Type} & \textbf{Updated element} & \textbf{Source reference} \\
\midrule
Variable
& $T \geq 0$: continuous bottleneck-throughput variable
& ``How should each worker be assigned to a station to maximize the overall production capacity of the assembly line?'' \\
Objective
& Maximize $T$
& ``How should each worker be assigned to a station to maximize the overall production capacity of the assembly line?'' \\
Constraint
& $T \leq \sum_{w} e_{w,s} \cdot x_{w,s}$ $\forall\, s$
& ``An assembly line has 5 stations, each responsible for one of the five steps in the assembly of a certain product.'' \\
\bottomrule
\end{tabular}
}

\smallskip
\textbf{Confidence:} Parameters 95 $\mid$ Variables 100 $\mid$ Objective 100 $\mid$ Constraints 100 $\mid$ \textbf{Min score: 95/100}
\end{examplebox}

\noindent The max-min criterion therefore selects $M_{3}$ (min = 95) over $M_{2}$ (min = 90) and $M_{1}$ (min = 30). It yields the correct bottleneck throughput ($T^* = 5.0$). By contrast, executing $M_{1}$ would optimize the mis-specified sum-of-efficiencies objective and return 28.0, which is not meaningful for the problem.
Appendix~\ref{sec:case_studies} provides more details for this example and an additional solver-dispatch case study.
\subsection{Multi-Solver Dispatch} \label{sec:solver_dispatch}

Because different OR problem classes are best served by different solvers and modeling paradigms, COOPA does not commit to a single solver backend. As shown on the right of Figure~\ref{fig:architecture}, the solver-dispatch component classifies the selected model $M^*$ and routes it to one of four specialized optimizer agents (Table~\ref{tab:optimizer_agents}). The optimizer agent is not the solver itself; it is the agent layer that decides how to encode the model, calls the appropriate backend library or external solver, inspects the output, and retries when execution fails. Each agent therefore packages solver-specific prompts (Appendix~\ref{sec:appendix_prompts}), whitelisted tool access, and domain-appropriate libraries around a particular solver family.

Each optimizer agent operates within a \emph{Thought--Code--Observation} loop: it analyzes the model, generates backend-specific Python code, executes that code to call the target solver or library, inspects the output, and retries if errors occur. The general optimizer handles problems that do not require a dedicated solver package, such as simulation-based or custom numerical tasks.

This design reflects OR practice: different problem classes benefit from different solvers. Vehicle routing and scheduling fit OR-Tools better than Pyomo, while non-convex or multi-objective problems may require evolutionary algorithms rather than branch-and-bound. For example, a multi-objective portfolio problem requires a Pareto front, so COOPA routes it to the metaheuristic optimizer rather than forcing repeated single-objective scalarizations. 
This design is also \emph{scalable}: adding support for a new solver requires only defining a new optimizer agent, with no changes to the modeling stage. Fine-tuned models such as ORLM \cite{tang2024orlm} could likewise be integrated as specialized agents within the same architecture.

\begin{table}[h]
\centering
\caption{Specialized Optimizer Agents in COOPA.}
\label{tab:optimizer_agents}
\begin{tabular}{lll}
\toprule
\textbf{Optimizer Agent} & \textbf{Problem Type} & \textbf{Backend Invoked} \\
\midrule
Mathematical Optimizer & LP, MIP, nonlinear & Pyomo \\
Combinatorial Optimizer & Routing, assignment, scheduling & OR-Tools \\
Metaheuristic Optimizer & Non-convex, black-box & pymoo \\
General Optimizer & Numerical, simulation, uncategorized & Standard Python \\
\bottomrule
\end{tabular}
\end{table}

\section{Experiments} \label{sec:experiments}

We evaluate COOPA across three benchmarks, eight LLM backbones, and four baselines. 

\subsection{Experimental Setup} \label{sec:exp_setup}




\textbf{Datasets.} We use three deterministic optimization benchmarks. \textbf{ComplexLP} contains 211 LP and MILP problems from MAMO \cite{huang2024mamo}; although all are linear, they require correct variable typing and constraint modeling. \textbf{IndustryOR} contains 100 real-world industrial OR problems from ORLM \cite{tang2024orlm}, spanning eight industries, three difficulty levels, and five problem types (LP, IP, MIP, NLP, and others). \textbf{BWOR} contains 82 textbook problems from OR-LLM-Agent \cite{zhang2025or}, covering LP, IP, and MIP with no nonlinear instances; two of these are infeasible and have no ground-truth optimum, so we evaluate on the remaining 80. We exclude NLP4LP, NL4OPT, and EasyLP because frontier models already reach saturation-level accuracy ($>80\%$), limiting their value for method comparison.


\textbf{LLM backbones.} We evaluate eight backbones spanning proprietary and open-source, reasoning and non-reasoning models: GPT-5.2, GPT-5, GPT-4.1, o3, o4-mini, Gemini-3-Flash, Gemini-2.5-Flash, and Qwen3-30B-A3B (abbreviated as Qwen3-30B), letting us test performance variation over LLMs.



\textbf{Baselines.} We compare against four prior LLM-based OR systems. Chain-of-Experts (CoE) \cite{xiao2023chain} uses a conductor model to orchestrate specialized expert agents and applies backward reflection after failure. OptiMUS \cite{ahmaditeshnizi2024optimus} decomposes problems into clauses and uses an evaluate-improve/debug loop with a Gurobi backend. OptiTree \cite{liu2025optitree} retrieves modeling thoughts from a hierarchical modeling tree. OR-LLM-Agent \cite{zhang2025or} uses a simple three-agent pipeline (Math $\to$ Code $\to$ Debugging), with correction driven mainly by downstream execution failures.


\textbf{Metric.} We report accuracy, defined as the fraction of problems whose reported objective value is within $0.1$ absolute error of the ground-truth optimum. Two BWOR problems are infeasible and excluded. We report macro-averages across the three benchmarks. 

\subsection{Main Results and Analysis} \label{sec:main_results}

The results are reported in Table~\ref{tab:main_results}. 
COOPA achieves the highest cross-model macro-average accuracy at 64.8\%, compared with 61.6\% for OR-LLM-Agent, 61.3\% for OptiTree, and 60.1\% for CoE. It achieves the best macro-average on 6 of 8 backbones. The 3.2-point gain over the next-best method is modest in absolute terms but consistent across multiple backbones.

\begin{table*}[!t]
\centering
\caption{Accuracy (\%) across three benchmarks and eight LLM backbones. \textbf{Bold} indicates best performance for each model. COOPA achieves the highest macro-average on 6 of 8 backbones.}
\label{tab:main_results}

{\footnotesize
\setlength{\tabcolsep}{5pt}
\renewcommand{\arraystretch}{0.88}
\begin{tabular}{ll cccc}
\toprule
\textbf{Model} & \textbf{Method} & \textbf{ComplexLP} & \textbf{IndustryOR} & \textbf{BWOR} & \textbf{Macro-Avg} \\
\midrule

\multirow{5}{*}{\textbf{GPT-5.2}}
 & Chain-of-Experts & 55.5 & 70.0 & 75.0 & 66.8 \\
 & OptiMUS & 14.2 & 14.0 & 25.0 & 17.7 \\
 & OptiTree & 53.6 & 74.0 & 77.5 & 68.4 \\
 & OR-LLM-Agent & 49.8 & 70.0 & 78.8 & 66.2 \\
 & \textbf{COOPA (Ours)} & \textbf{55.9} & \textbf{76.0} & \textbf{80.0} & \textbf{70.6} \\

\midrule

\multirow{5}{*}{\textbf{GPT-5}}
 & Chain-of-Experts & 48.8 & 63.0 & 76.3 & 62.7 \\
 & OptiMUS & 38.4 & 43.0 & 47.5 & 43.0 \\
 & OptiTree & 43.1 & 54.0 & 52.5 & 49.9 \\
 & OR-LLM-Agent & 40.3 & 63.0 & 67.5 & 56.9 \\
 & \textbf{COOPA (Ours)} & \textbf{53.1} & \textbf{75.0} & \textbf{80.0} & \textbf{69.4} \\

\midrule

\multirow{5}{*}{\textbf{GPT-4.1}}
 & Chain-of-Experts & 43.6 & 60.0 & 70.0 & 57.9 \\
 & OptiMUS & 39.3 & 48.0 & 58.8 & 48.7 \\
 & OptiTree & 49.3 & 62.0 & 68.8 & 60.0 \\
 & OR-LLM-Agent & 45.5 & 61.0 & 71.3 & 59.3 \\
 & \textbf{COOPA (Ours)} & \textbf{53.6} & \textbf{69.0} & \textbf{76.3} & \textbf{66.3} \\

\midrule

\multirow{5}{*}{\textbf{o3}}
 & Chain-of-Experts & \textbf{55.5} & 72.0 & 75.0 & \textbf{67.5} \\
 & OptiMUS & 36.0 & 43.0 & 42.5 & 40.5 \\
 & OptiTree & 43.1 & \textbf{75.0} & 78.8 & 65.6 \\
 & OR-LLM-Agent & 47.9 & 64.0 & \textbf{80.0} & 64.0 \\
 & \textbf{COOPA (Ours)} & 53.6 & 73.0 & 73.8 & 66.8 \\

\midrule

\multirow{5}{*}{\textbf{o4-mini}}
 & Chain-of-Experts & 48.3 & 69.0 & 68.8 & 62.0 \\
 & OptiMUS & 34.6 & 43.0 & 43.8 & 40.5 \\
 & OptiTree & \textbf{52.6} & 65.0 & 71.3 & 63.0 \\
 & OR-LLM-Agent & 47.4 & 68.0 & \textbf{78.8} & 64.7 \\
 & \textbf{COOPA (Ours)} & 47.9 & \textbf{72.0} & 77.5 & \textbf{65.8} \\

\midrule

\multirow{5}{*}{\textbf{Gemini-3-Flash}}
 & Chain-of-Experts & 47.4 & \textbf{75.0} & 75.0 & 65.8 \\
 & OptiMUS & 38.4 & 26.0 & 28.8 & 31.1 \\
 & OptiTree & \textbf{60.2} & 67.0 & 75.0 & 67.4 \\
 & OR-LLM-Agent & 52.6 & 69.0 & \textbf{81.3} & 67.6 \\
 & \textbf{COOPA (Ours)} & 52.6 & \textbf{75.0} & 77.5 & \textbf{68.4} \\

\midrule

\multirow{5}{*}{\textbf{Gemini-2.5-Flash}}
 & Chain-of-Experts & 49.3 & 32.0 & 47.5 & 42.9 \\
 & OptiMUS & 27.0 & 35.0 & 36.3 & 32.8 \\
 & OptiTree & \textbf{53.6} & 62.0 & 70.0 & 61.9 \\
 & OR-LLM-Agent & 46.4 & 67.0 & 70.0 & 61.1 \\
 & \textbf{COOPA (Ours)} & 47.4 & \textbf{71.0} & \textbf{77.5} & \textbf{65.3} \\

\midrule

\multirow{5}{*}{\textbf{Qwen3-30B\textsuperscript{\dag}}}
 & Chain-of-Experts & 42.2 & 57.0 & \textbf{67.5} & \textbf{55.6} \\
 & OptiMUS & 23.7 & 28.0 & 35.0 & 28.9 \\
 & OptiTree & \textbf{46.4} & 55.0 & 61.3 & 54.2 \\
 & OR-LLM-Agent & 38.9 & \textbf{58.0} & 62.5 & 53.1 \\
 & \textbf{COOPA (Ours)} & 32.2 & 48.0 & 56.3 & 45.5 \\

\bottomrule
\end{tabular}
}
\end{table*}
\begingroup
\renewcommand{\thefootnote}{\dag}%
\footnotetext{Qwen3-30B is shorthand for Qwen3-30B-A3B-Thinking-2507.}%
\endgroup

\textbf{Gains vary with backbone capability.} COOPA's largest gains over the best baseline appear on GPT-5 (+6.7 points: 69.4 vs.\ 62.7 for CoE) and GPT-4.1 (+6.3: 66.3 vs.\ 60.0 for OptiTree). Gains are smaller on GPT-5.2 (+2.2) and on the reasoning models o3 ($-$0.7, where CoE leads) and o4-mini (+1.1). This pattern suggests that iterative refinement helps most when the backbone is strong enough to benefit from self-correction but not already near saturation.

\textbf{IndustryOR advantage.} COOPA is especially strong on IndustryOR, the most ambiguous benchmark. On GPT-5 and Gemini-3-Flash, it reaches 75.0\%; on Gemini-3-Flash this ties the best baseline, while on GPT-5 it is the top result. This is consistent with our motivating claim that COOPA is most useful when the main bottleneck is problem-to-model abstraction rather than code execution.

\textbf{No single baseline dominates.} The strongest baseline changes across backbones: CoE leads on o3 and Qwen3-30B, OptiTree is best among baselines on GPT-5.2, and OR-LLM-Agent is strongest on Gemini-3-Flash. This reinforces the need for cross-model evaluation rather than single-model benchmarking.

\subsection{Additional Analysis Summary}



Appendix~\ref{sec:appendix_experiments} reports the supplementary experiment results. Here we retain only the two findings most central to the main claim.

\textbf{Ablation.} Iterative confidence-based modeling is the main driver of COOPA's gains. Relative to solving only the first candidate from the same run, the full $k=3$ pipeline improves macro-average accuracy on 7 of 8 backbones and raises the cross-model mean from 61.8\% to 64.8\% (+3.0 points). Without iteration, the base pipeline is roughly tied with the strongest baselines, indicating that iterative refinement explains most of COOPA's advantage.

\textbf{Confidence analysis.} The max-min confidence criterion adds signal beyond candidate diversity. Solving the selected candidate instead of the first candidate improves accuracy on 7 of 8 backbones, confidence gain correlates positively with accuracy gain across model--benchmark pairs ($r = 0.58$, $p = 0.003$), and the criterion makes 181 beneficial overrides versus 95 harmful ones. The appendix also provides full results on cross-model robustness, the Qwen3-30B underperformance case, solver-dispatch statistics, cost, and case studies.

\section{Related Work} \label{sec:related_work}

The literature on LLM-based OR has grown rapidly. We organize prior work into three categories: workflow and agent approaches that use general-purpose LLMs, training-based approaches that adapt model weights, and methodological foundations in self-improvement and multi-agent systems. Table~\ref{tab:design_comparison} summarizes how existing agent-based methods compare along key design dimensions.

\subsection{Workflow and Agent Approaches} \label{sec:rw_pipeline}

Early work on LLM-based OR workflows established two complementary strategies for managing problem complexity. \textbf{OptiMUS} \cite{ahmaditeshnizi2024optimus} introduces a modular workflow that decomposes problems into individual clauses, formulates each into mathematical expressions independently, and uses a connection graph to manage context across clauses. Importantly, OptiMUS is the first system to incorporate formulation quality assessment: a confidence-based feedback mechanism scores each clause on a 1--5 scale and can escalate uncertain formulations to a stronger model. \textbf{Chain-of-Experts (CoE)} \cite{xiao2023chain} takes a different approach by orchestrating 11 specialized agents (e.g., Terminology Interpreter, Variable Extraction, Code Reviewer) through a Conductor LLM that dynamically selects which expert to invoke. When the initial solution fails, a backward reflection mechanism re-consults experts in reverse order to identify and correct errors. While OptiMUS focuses on decomposing the \emph{problem} into manageable pieces, CoE focuses on decomposing the \emph{workflow} into specialized roles.

To address the difficulty of formulating problems from scratch, subsequent methods incorporate external knowledge into the modeling process. \textbf{OptiTree} \cite{liu2025optitree} constructs a hierarchical modeling tree offline, organized by problem taxonomy and complexity, so that at inference time the system can retrieve modeling thoughts from the closest known subproblem as context for formulation. This knowledge-augmented approach is particularly effective for problems that fall within the tree's coverage, though it generates formulations in a single pass at inference time. \textbf{LEAN-LLM-OPT} \cite{liang2026llm} addresses a complementary challenge: large-scale problems where input data resides in external files and exceeds prompt length limits. It uses RAG-based retrieval from 96 reference examples to classify problems and dynamically construct step-by-step workflows, offloading mechanical data-handling operations to auxiliary tools. Both methods demonstrate the value of grounding LLM generations in external knowledge, but neither assesses formulation quality before code generation.

Other methods explore different design priorities. \textbf{OR-LLM-Agent} \cite{zhang2025or} argues that modern reasoning LLMs have sufficiently strong mathematical capabilities to benefit from simple task decomposition rather than elaborate prompting or retrieval. It employs three sequential agents, a Math Agent, a Code Agent, and a Debugging Agent, achieving competitive results with minimal workflow complexity. \textbf{OptimAI} \cite{thind2025optimai} stands out for supporting multiple solver backends (including PuLP, Pyomo, and OR-Tools), broadening the range of problems that can be addressed. It generates multiple solution plans and falls back to alternatives when the initial plan fails.

\begin{table}[t]
\centering
\caption{Design comparison of agent-based LLM-OR methods. \emph{External knowledge}: whether the system retrieves or references pre-built examples or taxonomies. \emph{Formulation quality assessment}: whether the system evaluates formulation correctness before code generation. \emph{Source traceability}: whether users can trace modeling elements back to the problem description. \emph{Multi-solver}: whether the system supports more than one solver backend.}
\label{tab:design_comparison}
\begin{tabular}{lcccc}
\toprule
\textbf{Method} & \makecell{\textbf{External}\\\textbf{Knowledge}} & \makecell{\textbf{Formulation Quality}\\\textbf{Assessment}} & \makecell{\textbf{Source}\\\textbf{Traceability}} & \textbf{Multi-Solver} \\
\midrule
OptiMUS & \xmark & \cmark & \xmark & \xmark \\
Chain-of-Experts & \xmark & \xmark & \xmark & \xmark \\
OptiTree & \cmark & \xmark & \xmark & \xmark \\
OR-LLM-Agent & \xmark & \xmark & \xmark & \xmark \\
LEAN-LLM-OPT & \cmark & \xmark & \xmark & \xmark \\
OptimAI & \xmark & \xmark & \xmark & \cmark \\
\midrule
\textbf{COOPA (ours)} & \xmark & \cmark & \cmark & \cmark \\
\bottomrule
\end{tabular}
\end{table}

Across these methods, two gaps remain. First, although OptiMUS introduces confidence scoring, it neither explains in natural language why a score is low nor uses evaluation to iteratively improve formulations. Other workflow methods either omit formulation-quality assessment or perform it reactively, after downstream execution failures rather than through direct analysis of the formulation. As a result, these systems fail to exploit LLMs’ demonstrated capacity for self-evaluation \cite{kadavath2022language, madaan2023selfrefine} to proactively refine mathematical formulations before code generation, leaving the dominant error source unaddressed.
Second, no existing method provides source traceability from individual modeling elements—parameters, variables, constraints, and objectives—to the specific problem text that motivates them. In practical OR problems with dozens or hundreds of constraints, this forces practitioners to manually reread the full problem description and reconstruct the LLM’s reasoning to verify correctness or locate errors, which does not scale.
COOPA addresses both gaps through iterative confidence-based modeling with four-dimensional scoring and natural-language explanations, plus source references that create a transparent audit trail from problem text to mathematical model.

\subsection{Training-Based Approaches} \label{sec:rw_training}

A parallel line of work improves OR modeling accuracy by adapting model weights rather than designing workflows. \textbf{ORLM} \cite{tang2024orlm} introduces OR-Instruct, a semi-automated workflow for synthesizing optimization training data from 686 real-world seed cases, and demonstrates that fine-tuned 7B-parameter models can outperform standard GPT-4 on several benchmarks. Building on the idea of a universal intermediate representation, \textbf{LLMOPT} \cite{jiang2024llmopt} proposes a five-element formulation (Sets, Parameters, Variables, Objective, Constraints) combined with multi-instruction supervised fine-tuning and KTO alignment. LLMOPT also introduces an auto-testing self-correction loop of up to 12 iterations, though the correction is driven by execution errors and solver logs rather than formulation quality assessment, so correctly-executing but mathematically-incorrect models can persist. Finally, DPLM \cite{zhou2025auto} addresses the unique formulation challenges of dynamic programming via a synthetic data-generation pipeline (DualReflect) designed to capture complex state transitions and recursive relationships.

More recent work applies reinforcement learning to further improve modeling accuracy. \textbf{OR-R1} \cite{ding2025or} uses test-time reinforcement learning (TGRPO) with majority voting across generated candidates as a pseudo-label, achieving strong results with as few as 100 SFT training samples. \textbf{SIRL} \cite{chen2025solver} takes a different approach by using the solver itself as a verifier to provide reward signals during training, incorporating domain-specific feedback directly into the learning process. These training-based methods are complementary to workflow approaches: COOPA's architecture could use fine-tuned models as backbone LLMs, potentially combining the benefits of both paradigms.

\subsection{Self-Improvement and Multi-Agent Systems} \label{sec:rw_foundations}

LLMs can iteratively improve their outputs through self-generated feedback. Self-Refine \cite{madaan2023selfrefine} demonstrates this across tasks such as code optimization and mathematical reasoning, where a model critiques and revises a single output over multiple rounds without external supervision. Kadavath et al.\ \cite{kadavath2022language} show that LLMs exhibit meaningful, though imperfectly calibrated, self-knowledge about their capabilities and can often distinguish correct from incorrect answers. In code generation, Self-Debug \cite{chen2024teaching} and Reflexion \cite{shinn2023reflexion} extend self-improvement with execution feedback and episodic memory, respectively. Across these methods, the model iterates on a single output by repeatedly revising its latest attempt.

COOPA departs from this single-output paradigm. Instead of revising one formulation in place, it generates $k$ independent candidates, each informed by confidence evaluations of all prior candidates, then selects the candidate with the highest minimum score across the four modeling dimensions. This separates \emph{generation} from \emph{selection}: later candidates can address earlier weaknesses, while final selection can fall back to an earlier candidate if a later revision introduces new errors.

\emph{Multi-agent architectures} decompose complex problems into specialized subtasks handled by dedicated agents. In OR, CoE \cite{xiao2023chain} shows the value of specialization by assigning workflow steps—terminology interpretation, variable extraction, and code review—to different agents. COOPA specializes agents along a different axis: each optimizer agent targets a problem class and solver, routing problems to the appropriate solving paradigm rather than forcing them into a single formulation.


\section{Conclusion} \label{sec:conclusion}

We proposed COOPA, a modular LLM agent architecture for OR modeling that combines three components: iterative confidence-based modeling with max-min selection, source traceability with confidence explanations, and solver dispatch to specialized optimizer agents. Together, these components target three practical weaknesses of prior systems: weak formulation refinement, opaque outputs, and reliance on a single solver.

Experiments across three benchmarks and eight LLM backbones show that COOPA achieves the highest cross-model average accuracy (64.8\%) and the best macro-average on 6 of 8 backbones. The main empirical driver is iterative modeling: compared with solving only the first candidate, the full pipeline improves accuracy on 7 of 8 backbones and raises the cross-model mean by 3.0 points. The interpretability features provide an audit trail from problem text to formulation, while the appendix presents additional evidence and case studies for confidence analysis and solver dispatch.

COOPA's gains are largest on GPT-5 (+6.7 points) and GPT-4.1 (+6.3), and remain consistent enough to produce a 3.2-point average improvement over the next-best method. This pattern suggests that structured refinement is most useful when the backbone is strong enough to benefit from self-correction but not already saturated.

Future work should test COOPA on broader OR benchmarks with greater problem-type diversity, evaluate whether traceability and confidence explanations improve human verification in user studies, and combine the architecture with training-based OR models such as ORLM or SIRL.

\setlength{\bibsep}{3pt plus 0.3ex}
\bibliographystyle{ims}
\bibliography{references}

\clearpage
\appendix
\section{System Prompts} \label{sec:appendix_prompts}

This section presents the core system prompts used by each agent in COOPA. We show the key instruction sections; boilerplate template variables (tool lists, managed agent lists, planning prompts) are omitted for brevity.

\subsection{Formulation Extraction Agent}

The formulation extraction agent maps a natural-language description to a structured \texttt{Optimization\-Formulation} schema. The system prompt is:

\begin{examplebox}{Formulation Extraction System Prompt}
\small
\begin{verbatim}
You are an operations-research formulation assistant.
Given a natural-language optimization problem you will
populate the OptimizationFormulation schema exactly.

Guidelines:
- Copy the entire user prompt verbatim into the
  `question` field.
- Enumerate every numeric fact or named constant inside
  `parameters`. Use meaningful names (e.g., cost_harry)
  and include a SourceReference with the exact quote
  and contextual note.
- Decision variables must capture domains precisely
  (binary/integer/continuous, bounds, logical
  implications). Use SourceReference entries quoting
  the sentence that motivated the variable or domain.
- The `objective.expression` should be an algebraic
  description that references variable names, and
  `variables_involved` must list those variable
  identifiers.
- Each constraint gets its own entry. Use algebraic
  expressions when possible; fall back to `logical`
  for implications. Every constraint needs a
  SourceReference quoting the relevant requirement.
Return valid JSON only. Do not add fields.
\end{verbatim}
\end{examplebox}

\subsection{Confidence Evaluator}

After each formulation extraction, a confidence evaluator scores the formulation on four dimensions (parameters, decision variables, objective, constraints) on a 0--100 scale. The evaluation prompt is:

\begin{examplebox}{Confidence Evaluation Prompt (Template)}
\small
\begin{verbatim}
You are an expert in optimization and mathematical
modeling. Your task is to evaluate the quality and
correctness of an optimization problem formulation.

Given:
1. **Raw Question**: {raw_question}
2. **Proposed Formulation**: {formulation_str}

Please evaluate the confidence (0-100) for each of
the following components:

1. **PARAMETERS**: Are all necessary parameters
   identified with correct values and units?
2. **DECISION VARIABLES**: Are all decision variables
   properly defined with correct domains?
3. **OBJECTIVE**: Is the objective function correct and
   does it properly represent what should be optimized?
4. **CONSTRAINTS**: Are all necessary constraints
   included and correctly formulated?

For each component, provide a confidence score from
0-100 and a brief explanation (1-3 sentences).
\end{verbatim}
\end{examplebox}

\subsection{Refinement Prompt}

After each iteration, the next candidate is refined using all previous iterations as context. In the default setting, the system completes all $k$ iterations rather than stopping at a confidence threshold:

\begin{examplebox}{Refinement Prompt (Template)}
\small
\begin{verbatim}
You are refining an optimization formulation. Review
all previous attempts and the feedback to create a
better formulation.

**Original Problem:** {raw_question}

**HISTORY OF PREVIOUS FORMULATIONS:**
--- Iteration 1 ---
Formulation: {past_formulation_str}
Confidence Scores:
- Parameters: {score}/100 - {explanation}
- Decision Variables: {score}/100 - {explanation}
- Objective: {score}/100 - {explanation}
- Constraints: {score}/100 - {explanation}
[... additional iterations ...]

Please create a REFINED formulation that addresses all
the identified issues from past iterations. Learn from
what worked well and avoid repeating mistakes. Pay
special attention to the components with lower
confidence scores.
\end{verbatim}
\end{examplebox}

\subsection{User Prompt for Solver Dispatch}

After iterative refinement selects the best formulation, it is formatted into a structured prompt and sent to the solver-dispatch component as the user message:

\begin{examplebox}{Formatted User Prompt for Solver Dispatch (Template)}
\small
\begin{verbatim}
Delegate the following operations research problem to
the correct optimizer agent:

## PARAMETERS:
- {name} ({type}): {description} = {value} [{units}]
...

## DECISION VARIABLES:
- {name} ({type}): {description} | Domain: {domain}
...

## OBJECTIVE:
- Sense: MAXIMIZE/MINIMIZE
- Description: {description}
- Expression: {expression}
- Variables involved: {variables}

## CONSTRAINTS:
1. {name} ({sense}):
   Expression: {expression}
   Variables: {variables}
...

## CRITICAL INSTRUCTIONS:
- Your role is solver dispatch. You MUST NOT solve this problem
  yourself.
- Your ONLY job is to delegate the COMPLETE problem
  above to the appropriate optimizer agent in your
  FIRST Code block.
- The optimizer agent will handle everything: saving
  parameters to JSON, building the solver, executing
  it, and returning the result.
- Do NOT call final_answer() in the same response where
  you call an optimizer agent. You MUST wait for the
  system to return the optimizer's REAL result first.
\end{verbatim}
\end{examplebox}

\subsection{Solver-Dispatch Prompt}

This component delegates the problem to the appropriate optimizer agent. Its system prompt defines the routing logic:

\begin{examplebox}{Solver-Dispatch System Prompt (core instructions)}
\small
\begin{verbatim}
You are responsible only for solver dispatch in an advanced multi-agent
operations research system. Your role is to orchestrate
specialized agents and tools to deliver correct, clear,
and actionable solutions.

=== CORE CONSTRAINTS ===
1. Immediately delegate the problem to an optimizer
   agent in your FIRST Code block.
2. Do not try to solve the problem directly. No
   internal reasoning, solver code, or calculations.
3. Do not iterate - Let optimizer agents handle all
   solution refinement.
4. Identify agent type and pass complete problem
   statement to chosen agent.

=== PROCEDURE ===
1. Clarify the Problem.
2. Select and Delegate to the Appropriate Optimizer:
   - mathematical_optimizer_agent for algebraic LP, MILP,
     and continuous NLP models.
   - combinatorial_optimizer_agent for routing,
     scheduling, CP-SAT, and other discrete problems
     best expressed in OR-Tools.
   - metaheuristic_optimizer_agent for metaheuristic
     or black-box search, especially multi-objective
     or non-convex cases.
   - general_optimizer_agent for simulation-based,
     custom algorithmic, or general scripting.
3. Review and Present Results.
4. Call final_answer with the final result.
\end{verbatim}
\end{examplebox}

\subsection{Mathematical Optimizer Agent}

Solves algebraic LP, MILP, and continuous NLP problems using Pyomo with GLPK and IPOPT.

\begin{examplebox}{Mathematical Optimizer System Prompt (core instructions)}
\small
\begin{verbatim}
You are an expert operations research assistant who
models and solves mathematical optimization problems
using Pyomo in Python, supported by solvers such as
GLPK and IPOPT.

=== SCOPE AND CAPABILITIES ===
Supported problem types:
- Linear Programming (LP)
- Mixed-Integer Linear Programming (MIP)
- Nonlinear Programming (NLP)

Default solver support:
- GLPK for LP/MIP
- IPOPT for continuous NLP

If a problem requires unsupported MINLP capabilities,
do not claim IPOPT can solve it directly. Either reformulate
to a supported class or report the limitation.

=== PROCEDURE ===
1. Understand the Problem.
2. Create Parameters JSON File.
3. Build the Solver with Pyomo and Save to Python File.
4. Load and Execute the Solver via
   load_object_from_python_file().
5. If execution FAILED: fix and re-execute.
6. Final Answer (only after successful execution).
\end{verbatim}
\end{examplebox}

\subsection{Combinatorial Optimizer Agent}

Solves VRP, scheduling, assignment, constraint satisfaction, and bin packing problems using Google OR-Tools.

\begin{examplebox}{Combinatorial Optimizer System Prompt (core instructions)}
\small
\begin{verbatim}
You are an expert combinatorial optimization assistant
specializing in solving combinatorial, routing, and
constraint satisfaction problems using Google OR-Tools.

=== SCOPE AND CAPABILITIES ===
- Vehicle Routing Problems (VRP, CVRP, VRPTW)
- Job Shop / Flow Shop Scheduling
- Assignment Problems
- Constraint Programming (CP-SAT)
- Bin Packing & Knapsack
- Graph traversal and network design

=== PROCEDURE ===
1. Understand the Problem.
2. Create Parameters JSON File.
3. Build the Solver with OR-Tools, Save to Python File.
4. Load and Execute via load_object_from_python_file().
5. If execution FAILED: fix and re-execute.
6. Final Answer (only after successful execution).
\end{verbatim}
\end{examplebox}

\subsection{Metaheuristic Optimizer Agent}

Solves multi-objective, non-convex, and black-box optimization problems using pymoo.

\begin{examplebox}{Metaheuristic Optimizer System Prompt (core instructions)}
\small
\begin{verbatim}
You are an expert meta-heuristic optimization assistant
specializing in solving complex, non-convex, and
multi-objective optimization problems using evolutionary
and meta-heuristic algorithms implemented in pymoo.

=== SCOPE AND CAPABILITIES ===
- Black-box or simulation-based objective functions
- Multi-objective scenarios (Pareto-optimal solutions)
- Non-convex, discontinuous, nonlinear landscapes
- Mixed-variable types (continuous, discrete, binary,
  permutations)

Algorithms available through pymoo:
- Single-objective: GA, DE, PSO, CMA-ES
- Multi-objective: NSGA-II, NSGA-III, MOEA/D

=== PROCEDURE ===
(Same as mathematical optimizer, using pymoo instead)
\end{verbatim}
\end{examplebox}

\subsection{General Optimizer Agent}

Handles simulation-based, custom algorithmic, or scripting tasks that do not fit the other categories.

\begin{examplebox}{General Optimizer System Prompt (core instructions)}
\small
\begin{verbatim}
You are the general purpose optimizer agent for
operations research and optimization. Your job is to
solve problems that do not fit into mathematical,
combinatorial, or metaheuristic categories. You are
especially suited for simulation-based, custom
algorithmic, or scripting tasks that require flexible
Python code.

=== SCOPE AND CAPABILITIES ===
You can write, modify, and execute Python code to solve
a wide range of operations research and simulation
problems. You are not limited to any specific
optimization paradigm.
\end{verbatim}
\end{examplebox}

\subsection{Shared Execution Constraints}

All four optimizer agents share the following critical execution rules, designed to prevent hallucinated solver outputs:

\begin{examplebox}{Shared Execution Rules (all optimizer agents)}
\small
\begin{verbatim}
=== CRITICAL: STOPPING CRITERIA ===
YOU MUST STOP GENERATING TEXT IMMEDIATELY AFTER WRITING
A CODE BLOCK.
- Do NOT simulate or fabricate the execution output.
- Do NOT hallucinate solver results.
- Wait for the actual system to execute your code.

=== CRITICAL: SOLVER EXECUTION IS MANDATORY ===
YOU MUST NEVER:
- Output just numbers from reasoning/analysis.
- Skip the solver step or assume results without
  running code.
- Provide final answers without actual solver
  execution results.

YOU MUST ALWAYS:
- Write complete, executable Python code blocks.
- Actually execute the solver in your code.
- Extract results ONLY from actual solver output.
\end{verbatim}
\end{examplebox}

\section{Supplementary Case Studies}\label{sec:case_studies}

\subsection{Iterative Modeling Corrects a Conceptual Error} 

We return to the assembly line example from Section~\ref{sec:method} (BWOR Question~76, o4-mini backbone) and show the corresponding COOPA execution logs. The example illustrates how iterative confidence-based modeling catches and corrects a fundamental objective error.

\textbf{Problem.} Five workers (A--E) must be assigned to five stations (I--V), one per station. Because the product passes through the line sequentially, throughput is determined by the \emph{slowest} station. The correct model is therefore max-min: maximize the minimum station throughput. The gold answer is $T^* = 5.0$ pieces/min.

\textbf{Without iterative modeling.} The first candidate (Iteration~1) instead maximizes the \emph{sum} of worker efficiencies across all stations:
\begin{equation*}
  \max \sum_{w \in W} \sum_{s \in S} e_{w,s} \cdot x_{w,s}
\end{equation*}
subject to one-worker-per-station and one-station-per-worker assignment constraints. This is a valid assignment model but the wrong objective for an assembly line: maximizing the sum (= 28) ignores the bottleneck, so the reported value 28.0 does not match the target throughput of 5.0.

\textbf{With iterative modeling, Iteration 1.} The first candidate has the \emph{same} incorrect sum-of-efficiencies objective. The confidence evaluator assigns only 30/100 to the objective dimension, with the explanation:

\begin{quote}
\emph{``The formulation maximizes the sum of efficiencies, but assembly-line capacity is driven by the bottleneck (minimum station rate), so the objective is mis-specified.''}
\end{quote}

\noindent The other three dimensions score highly (parameters: 95, variables: 100, constraints: 100), so the evaluator isolates the objective as the only weak dimension. The candidate's minimum confidence is therefore 30/100.

\textbf{Iteration 2.} Guided by that low objective score, the LLM restructures the formulation around a bottleneck-throughput variable $T \geq 0$, replaces the objective with $\max\; T$, and adds bottleneck constraints $T \leq \sum_{w} e_{w,s} \cdot x_{w,s}$ for each station $s$. This is the correct max-min formulation, and its minimum confidence rises to 90/100.

\textbf{Iteration 3 and selection.} The third candidate keeps the same bottleneck structure with minor refinements and reaches the highest minimum confidence, 95/100. Max-min selection therefore chooses Iteration~3 over Iteration~2 (90) and Iteration~1 (30). The mathematical optimizer then returns the correct bottleneck throughput, $T^* = 5.0$.

\textbf{Analysis.} This case highlights three points. First, the confidence evaluator catches a \emph{conceptual} error, not just a syntactic one. Second, the fix is structural: Iteration~2 introduces a new decision variable $T$ and bottleneck constraints rather than patching a local mistake. Third, without iterative modeling, COOPA returns the same wrong answer (28.0), showing that the gain comes from the confidence-and-refinement loop itself. Table~\ref{tab:case_study_summary} summarizes the progression.

\begin{table}[h]
\centering
\caption{Case study summary: confidence scores and outcomes across three iterations for BWOR Question~76 (o4-mini).}
\label{tab:case_study_summary}
\begin{tabular}{lccccccc}
\toprule
& \textbf{Params} & \textbf{Vars} & \textbf{Obj} & \textbf{Constrs} & \textbf{Min} & \textbf{Objective type} & \textbf{Answer} \\
\midrule
Iteration 1 & 95 & 100 & 30 & 100 & 30 & Sum (wrong) & 28.0 \\
Iteration 2 & 90 & 90 & 100 & 95 & 90 & Max-min (correct) & 5.0 \\
Iteration 3 & 95 & 100 & 100 & 100 & \textbf{95} & Max-min (correct) & 5.0 \\
\midrule
\multicolumn{6}{l}{w/o iterative modeling} & Sum (wrong) & 28.0 \\
\bottomrule
\end{tabular}
\end{table}


\begin{examplebox}{Solver Dispatch and Result}
\small
\textbf{Routing decision:} Mixed-integer program (binary assignment + continuous $T$).

\textbf{Routed to:} mathematical optimizer (Pyomo).

\medskip
\textbf{Generated code:}
\begin{verbatim}
model = ConcreteModel()
model.WORKERS = Set(initialize=params["workers"])
model.STATIONS = Set(initialize=params["stations"])
model.x = Var(model.WORKERS, model.STATIONS, domain=Binary)
model.T = Var(domain=NonNegativeReals)

def one_station_per_worker_rule(m, w):
  return sum(m.x[w, s] for s in m.STATIONS) == 1
model.one_station_per_worker = Constraint(
  model.WORKERS, rule=one_station_per_worker_rule)

def one_worker_per_station_rule(m, s):
  return sum(m.x[w, s] for w in m.WORKERS) == 1
model.one_worker_per_station = Constraint(
  model.STATIONS, rule=one_worker_per_station_rule)

def bottleneck_rule(m, s):
  return m.T <= sum(
    params["efficiency"][w][s] * m.x[w, s]
    for w in m.WORKERS)
model.bottleneck = Constraint(
  model.STATIONS, rule=bottleneck_rule)

model.obj = Objective(expr=model.T, sense=maximize)
solver = SolverFactory('glpk')
result = solver.solve(model)
\end{verbatim}

\textbf{Execution result:}
\begin{verbatim}
status: optimal
objective: 5.0
assignment: A->V, B->IV, C->II, D->I, E->III
T: 5.0
\end{verbatim}

\emph{Without iterative modeling, the incorrect formulation optimizes $\sum e_{w,s}$ and returns 28.0, which is not the target throughput.}
\end{examplebox}

\subsection{Multi-Objective Optimization via Solver Dispatch} \label{sec:appendix_car_side_impact}

To demonstrate the value of COOPA's multi-solver dispatch, we present a case study on the car side impact design problem \cite{jain2014evolutionary}, a well-known multi-objective benchmark from the engineering optimization literature. This problem has 7 continuous decision variables (structural thicknesses), 3 conflicting objectives (minimize weight, minimize pubic symphysis force, minimize average V-pillar velocity), and 10 nonlinear safety constraints including bilinear and quadratic terms. The goal is to find the \emph{Pareto front}: the set of non-dominated designs representing the best tradeoffs among the three objectives.

This problem poses two challenges for systems locked to a single solver. First, it is \emph{multi-objective}: the designer needs the full Pareto front, not a single optimal point, because the tradeoff between weight and safety depends on domain priorities. Gurobi and PuLP can optimize scalarized versions of the problem, but they do not directly return a Pareto front; obtaining one requires manual scalarization or repeated $\epsilon$-constraint solves. Second, the constraints contain \emph{bilinear terms} ($x_2 x_4$, $x_1 x_2$, etc.) and \emph{quadratic terms} ($x_2^2$), which PuLP cannot handle and which require Gurobi's non-convex QCQP mode. We fed the same natural-language problem description to both COOPA and OR-LLM-Agent (the best baseline by cross-model mean accuracy) and compared their generated solutions.

\subsubsection{OR-LLM-Agent Solution (Gurobi)}

OR-LLM-Agent recognized that the problem is multi-objective and nonlinear. Its Math Agent correctly identified all three objectives, all constraints, and noted that the epsilon-constraint method is needed for multi-objective optimization with Gurobi. However, the system encountered two failures in sequence.

\textbf{Attempt 1: Epsilon-constraint (infeasible).} The LLM chose $\epsilon_F = 3.5$ and $\epsilon_V = 12.0$ as bounds for the force and velocity objectives, converting the multi-objective problem into a single-objective one (minimize weight). Gurobi reported the model as \emph{infeasible}: no design exists that simultaneously satisfies all 10 safety constraints and the epsilon bounds. The LLM selected these epsilon values without prior knowledge of the feasible objective space, and the chosen combination was too restrictive.

\textbf{Attempt 2: Self-repair (single objective only).} The Debugging Agent detected the infeasibility and self-repaired by removing the epsilon constraints entirely, reducing the problem to minimizing weight subject only to the safety constraints. Gurobi solved this simplified problem and returned a single feasible design:

\begin{examplebox}{{OR-LLM-Agent Result (Gurobi, after self-repair)}}
\small
\textbf{Approach:} Minimize weight only (abandoned multi-objective)

\medskip
\textbf{Result:} One feasible point (not a Pareto front)\\[3pt]
\begin{tabular}{@{}ll@{}}
Weight ($f_1$): & 23.5857 \\
Pubic force ($f_2$): & 4.0000 (at safety limit) \\
Avg V-velocity ($f_3$): & 12.5211 \\
\end{tabular}

\medskip
\textbf{Design:}\\[3pt]
\begin{tabular}{@{}ll@{}}
$x_1$ (B-pillar inner): & 0.5000 \quad $x_5$ (door beam): 0.8750 \\
$x_2$ (B-pillar reinforce): & 1.2257 \quad $x_6$ (door belt): 0.8843 \\
$x_3$ (floor side inner): & 0.5000 \quad $x_7$ (roof rail): 0.4000 \\
$x_4$ (cross member): & 1.2071 \\
\end{tabular}

\medskip
\textbf{Execution result:}
\begin{verbatim}
Attempt 1 (epsilon-constraint):
  Barrier solved model in 0 iterations and 0.00s
  Model is infeasible.
  IIS computed: 1 constraint, 4 bounds
Attempt 2 (self-repair, minimize weight only):
  Optimal solution found (tolerance 1.00e-04)
  Best objective 2.358565798676e+01, gap 0.0000%
  f1 (weight)      = 23.5857
  f2 (pubic force)  = 4.0000
  f3 (avg velocity) = 12.5211
  x = [0.5000, 1.2257, 0.5000, 1.2071,
       0.8750, 0.8843, 0.4000]
\end{verbatim}

\medskip
\textbf{Problem:} This is the minimum-weight extreme of the Pareto front. The force ($f_2 = 4.00$) is at its maximum safety limit, meaning passenger safety is sacrificed entirely for weight reduction. A decision-maker would need to see the full tradeoff surface to make an informed choice.
\end{examplebox}

The generated code is shown below. Note the 145-line implementation with 9 auxiliary variables for bilinear terms, explicit quadratic constraint definitions, and the epsilon-constraint workaround that ultimately had to be abandoned.

\begin{examplebox}{OR-LLM-Agent Generated Code (excerpts)}
\small
\begin{verbatim}
# 9 auxiliary variables for bilinear terms
q_12 = m.addVar(name="q_12")   # x1 * x2
q_23 = m.addVar(name="q_23")   # x2 * x3
q_24 = m.addVar(name="q_24")   # x2 * x4
q_26 = m.addVar(name="q_26")   # x2 * x6
q_27 = m.addVar(name="q_27")   # x2 * x7
q_35 = m.addVar(name="q_35")   # x3 * x5
q_37 = m.addVar(name="q_37")   # x3 * x7
q_56 = m.addVar(name="q_56")   # x5 * x6
q_2sq = m.addVar(name="q_2sq") # x2^2

# Quadratic equality constraints to define aux vars
m.addQConstr(q_12 == x[0] * x[1], "def_q_12")
m.addQConstr(q_23 == x[1] * x[2], "def_q_23")
...  # (9 total quadratic equalities)

# Epsilon-constraint (caused infeasibility)
m.addQConstr(F_PS <= 3.5, "c_epsilon_F_PS")
m.addQConstr(V_avg <= 12.0, "c_epsilon_V_avg")

# After self-repair: removed epsilon constraints,
# minimize weight only
m.setObjective(W, GRB.MINIMIZE)
m.params.NonConvex = 2
m.optimize()
\end{verbatim}
\end{examplebox}

\subsubsection{COOPA Solution (pymoo)}

COOPA's solver-dispatch component classified this as a multi-objective nonlinear problem and routed it to the metaheuristic optimizer, which used pymoo's NSGA-II algorithm. The generated code directly defines the problem class with all three objectives and 10 constraints, then runs the evolutionary algorithm to compute the full Pareto front.

\begin{examplebox}{COOPA Result (pymoo NSGA-II)}
\small
\textbf{Approach:} NSGA-II, all 3 objectives optimized simultaneously

\medskip
\textbf{Result:} Full Pareto front (set of non-dominated designs)\\[3pt]
\begin{tabular}{@{}ll@{}}
Weight ($f_1$): & $[23.61, 42.70]$ \\
Pubic force ($f_2$): & $[3.59, 4.00]$ \\
Avg V-velocity ($f_3$): & $[10.62, 12.44]$ \\
\end{tabular}

\medskip
\textbf{Execution result:}
\begin{verbatim}
Optimization completed successfully.
Pareto front: 100 non-dominated solutions found
  f1 (weight)      in [23.61, 42.70]
  f2 (pubic force)  in [3.59, 4.00]
  f3 (avg velocity) in [10.62, 12.44]
Sample Pareto-optimal designs:
  Design A: f1=23.61, f2=4.00, f3=12.43 (lightest)
  Design B: f1=31.52, f2=3.73, f3=11.68 (balanced)
  Design C: f1=42.70, f2=3.59, f3=10.62 (safest)
\end{verbatim}

\medskip
\textbf{Output:} A population of Pareto-optimal designs, each representing a different tradeoff. A decision-maker can select the design that best balances weight reduction against passenger safety.
\end{examplebox}

\begin{examplebox}{COOPA Generated Code (excerpts)}
\small
\begin{verbatim}
class CarStructureOptimization(Problem):
    def __init__(self, params):
        super().__init__(
            n_var=7, n_obj=3, n_constr=10,
            xl=np.array([...]),  # lower bounds
            xu=np.array([...])   # upper bounds
        )

    def _evaluate(self, X, out, *args, **kwargs):
        x1, x2, ..., x7 = X[:,0], X[:,1], ..., X[:,6]

        # Objectives (directly expressed)
        f1 = 1.98 + 4.9*x1 + 6.67*x2 + ...
        f2 = 4.72 - 0.5*x4 - 0.19*x2*x3
        f3 = 0.5 * (V_MBP + V_FD)
        out["F"] = np.column_stack([f1, f2, f3])

        # Constraints (directly expressed)
        g = np.zeros((X.shape[0], 10))
        g[:,0] = (1.16 - 0.3717*x2*x4 ...) - 1.0
        ...
        out["G"] = g

# Solve with NSGA-II
algorithm = NSGA2(pop_size=100)
res = minimize(problem, algorithm,
               termination=('n_gen', 200))
\end{verbatim}
\end{examplebox}

\subsubsection{Comparison}

\begin{figure*}[t]
\begin{minipage}[t]{0.48\textwidth}
\begin{examplebox}{OR-LLM-Agent (Gurobi)}
\small
\textbf{Solver:} Gurobi (non-convex QCQP)\\
\textbf{Multi-objective method:} Epsilon-constraint\\
\textbf{Code complexity:} 145 lines\\
\textbf{Aux variables needed:} 9 (for bilinear terms)\\
\textbf{Attempt 1:} Infeasible (bad $\epsilon$ values)\\
\textbf{Attempt 2:} Minimized weight only\\
\textbf{Output:} \colorbox{red!15}{\textbf{1 point}} (min-weight extreme)\\
\textbf{Force ($f_2$):} 4.00 (at safety limit)\\
\textbf{Pareto front:} Not computed
\end{examplebox}
\end{minipage}
\hfill
\begin{minipage}[t]{0.48\textwidth}
\begin{examplebox}{COOPA (pymoo)}
\small
\textbf{Solver:} pymoo NSGA-II\\
\textbf{Multi-objective method:} Native (evolutionary)\\
\textbf{Code complexity:} 50 lines\\
\textbf{Aux variables needed:} 0\\
\textbf{Attempt 1:} Succeeded directly\\
\phantom{\textbf{Attempt 2:}} \\
\textbf{Output:} \colorbox{green!15}{\textbf{Full Pareto front}}\\
\textbf{Force ($f_2$):} $[3.59, 4.00]$ (full range)\\
\textbf{Pareto front:} Complete tradeoff surface
\end{examplebox}
\end{minipage}
\end{figure*}

\textbf{Analysis.} The comparison reveals three advantages of COOPA's multi-solver dispatch for this problem class.

First, \emph{paradigm matching}: NSGA-II is designed for multi-objective optimization and computes the full Pareto front in a single run. By contrast, a Gurobi-based approach must rely on scalarization or repeated $\epsilon$-constraint solves, which requires choosing objective tradeoffs without knowing the feasible objective space. When the LLM's guesses are infeasible, the solver fails entirely.

Second, \emph{code simplicity}: pymoo's API lets the LLM express objectives and constraints as direct function evaluations (50 lines). Gurobi requires auxiliary variables for every bilinear term, quadratic constraint definitions, and manual epsilon-constraint setup (145 lines). Simpler code means fewer opportunities for the LLM to introduce errors.

Third, \emph{solution completeness}: even after self-repair, OR-LLM-Agent returns only one extreme point (minimum weight with force at the safety limit). A decision-maker cannot see the tradeoff between weight and safety. COOPA returns the full Pareto front, enabling informed design decisions.

We also tested CoE on this problem; its first attempt with $\epsilon$-constraint was similarly infeasible, and its second attempt returned the same single-point minimum-weight solution (weight = 23.5857), confirming that the limitation is not specific to one baseline, but arises more broadly when these pipelines attack the task through repeated scalarized solves.

\section{Supplementary Experiments} \label{sec:appendix_experiments}

\subsection{Cross-Model Consistency Analysis} \label{sec:cross_model}

To measure robustness across backbones, we summarize each method's macro-average accuracy over the eight models in Table~\ref{tab:cross_model}. COOPA has the highest mean accuracy (64.8\%) and highest maximum (70.6\%). OR-LLM-Agent has the lowest variance (5.0) but a lower mean, indicating steadier but weaker performance. COOPA's variance is driven mainly by Qwen3-30B. Overall, single-backbone evaluation can be misleading: the leader changes by model, whereas COOPA has the best mean with moderate variance.


\begin{table}[h]
  \centering
  \caption{Cross-model consistency: summary statistics of macro-average accuracy (\%) across 8 LLMs.}
  \label{tab:cross_model}
  \begin{tabular}{lcccc}
  \toprule
  \textbf{Method} & \textbf{Mean} & \textbf{Std} & \textbf{Min} & \textbf{Max} \\
  \midrule
  Chain-of-Experts & 60.1 & 8.1 & 42.9 & 67.5 \\
  OptiMUS & 35.4 & 9.8 & 17.7 & 48.7 \\
  OptiTree & 61.3 & 6.5 & 49.9 & 68.4 \\
  OR-LLM-Agent & 61.6 & 5.0 & 53.1 & 67.6 \\
  \textbf{COOPA (Ours)} & \textbf{64.8} & 7.5 & 45.5 & \textbf{70.6} \\
  \bottomrule
  \end{tabular}
\end{table}

\subsection{Ablation Study: Effect of Iterative Modeling} \label{sec:ablation}

To isolate iterative confidence-based modeling, we compare two configurations from the same run: (1)~solve only the first candidate (Iteration~1), and (2)~run the full $k=3$ pipeline with confidence evaluation and max-min selection. We extract the first candidate from the iterative logs and solve it independently, so both conditions share the same generation context and differ only in candidate selection. All other components remain fixed. We focus on this ablation because iterative modeling is the main algorithmic contribution; structured output mainly supports human verification (Section~\ref{sec:structured_modeling}), and solver dispatch has limited signal on the current benchmarks because 91.1\% of problems are routed to the mathematical optimizer (Section~\ref{sec:solver_stats}).

\begin{table}[t]
  \centering
  \caption{Ablation of iterative modeling. Both rows are drawn from the same iterative run: ``w/o iteration'' solves only the first candidate (Iteration~1); ``w/ iteration'' solves the candidate selected by the max-min confidence criterion among $k=3$ candidates. $\Delta$ is w/ iteration minus w/o iteration.}
  \label{tab:ablation_results}
  \setlength{\tabcolsep}{4pt}
  \begin{tabular}{llccccc}
  \toprule
  \textbf{Model} & \textbf{Configuration} & \textbf{ComplexLP} & \textbf{IndustryOR} & \textbf{BWOR} & \textbf{Macro-Avg} &
  $\boldsymbol{\Delta}$ \\
  \midrule
  \multirow{2}{*}{GPT-5.2}
   & w/o iteration & 52.6 & \textbf{77.0} & \textbf{82.5} & \textbf{70.7} & \\
   & w/ iteration & \textbf{55.9} & 76.0 & 80.0 & 70.6 & $-$0.1 \\
  \midrule
  \multirow{2}{*}{GPT-5}
   & w/o iteration & 52.1 & 74.0 & 77.5 & 67.9 & \\
   & w/ iteration & \textbf{53.1} & \textbf{75.0} & \textbf{80.0} & \textbf{69.4} & $+$1.5 \\
  \midrule
  \multirow{2}{*}{GPT-4.1}
   & w/o iteration & 51.7 & 67.0 & 71.3 & 63.3 & \\
   & w/ iteration & \textbf{53.6} & \textbf{69.0} & \textbf{76.3} & \textbf{66.3} & $+$3.0 \\
  \midrule
  \multirow{2}{*}{o3}
   & w/o iteration & 49.8 & 73.0 & 71.3 & 64.7 & \\
   & w/ iteration & \textbf{53.6} & 73.0 & \textbf{73.8} & \textbf{66.8} & $+$2.1 \\
  \midrule
  \multirow{2}{*}{o4-mini}
   & w/o iteration & 45.5 & \textbf{73.0} & 72.5 & 63.7 & \\
   & w/ iteration & \textbf{47.9} & 72.0 & \textbf{77.5} & \textbf{65.8} & $+$2.1 \\
  \midrule
  \multirow{2}{*}{Gemini-3-Flash}
   & w/o iteration & 50.7 & 75.0 & 76.3 & 67.3 & \\
   & w/ iteration & \textbf{52.6} & 75.0 & \textbf{77.5} & \textbf{68.4} & $+$1.1 \\
  \midrule
  \multirow{2}{*}{Gemini-2.5-Flash}
   & w/o iteration & 45.5 & 66.0 & 70.0 & 60.5 & \\
   & w/ iteration & \textbf{47.4} & \textbf{71.0} & \textbf{77.5} & \textbf{65.3} & $+$4.8 \\
  \midrule
  \multirow{2}{*}{Qwen3-30B}
   & w/o iteration & 28.9 & 42.0 & 38.8 & 36.6 & \\
   & w/ iteration & \textbf{32.2} & \textbf{48.0} & \textbf{56.3} & \textbf{45.5} & $+$8.9 \\
  \bottomrule
  \end{tabular}
\end{table}

\begin{figure*}[t]
  \centering
  \includegraphics[width=\textwidth]{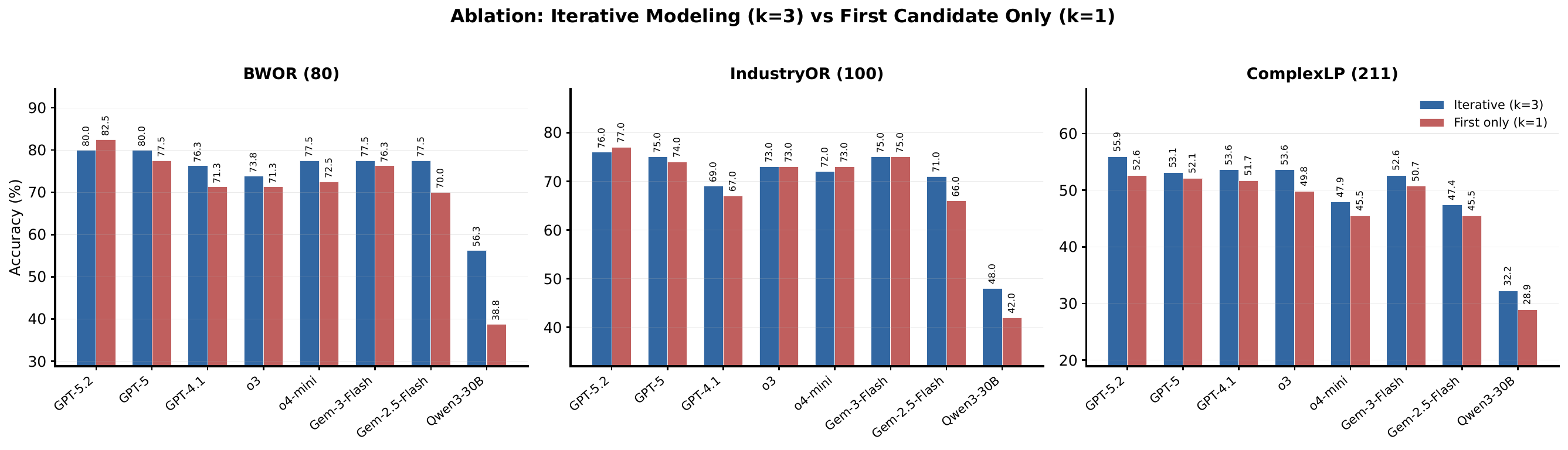}
  \caption{Per-benchmark ablation: iterative modeling ($k=3$, max-min selection) vs.\ first candidate only ($k=1$). Iterative modeling improves accuracy on the majority of backbone--benchmark pairs, with the largest gains on BWOR and ComplexLP.}
  \label{fig:ablation_3panel}
\end{figure*}

\textbf{Overall effect.} Table~\ref{tab:ablation_results} and Figure~\ref{fig:ablation_3panel} show gains on 7 of 8 backbones. The cross-model mean rises from 61.8\% to 64.8\% (+3.0 points). The largest improvements appear on Qwen3-30B (+8.9), Gemini-2.5-Flash (+4.8), and GPT-4.1 (+3.0). GPT-5.2 is essentially unchanged ($-$0.1).

\textbf{Gains are largest on weaker backbones.} Qwen3-30B gains the most (+8.9), while GPT-5.2 gains the least ($-$0.1). This suggests weaker models produce more correctable first-pass errors, leaving more room for confidence-based selection. The effect also appears on reasoning models such as o3 and o4-mini (+2.1 each), so it is not limited to non-reasoning architectures.

\textbf{Takeaway.} Iterative confidence-based modeling is the main driver of COOPA's advantage. Without it, the base pipeline reaches 61.8\%, roughly matching OR-LLM-Agent (61.6\%) and OptiTree (61.3\%). The extra 3.0 points from iteration explain most of COOPA's 3.2-point lead over the next-best baseline, consistent with the error analysis in Section~\ref{sec:modeling_errors}.

\subsection{Confidence Calibration Analysis} \label{sec:calibration}

The ablation shows that iterative modeling helps, but it leaves one question: does max-min confidence actually pick better candidates, or does it mainly benefit from generating more of them? To separate selection quality from candidate diversity, we analyze confidence scores and candidate-level correctness from COOPA's execution logs.

\begin{figure*}[t]
  \centering
  \includegraphics[width=\textwidth]{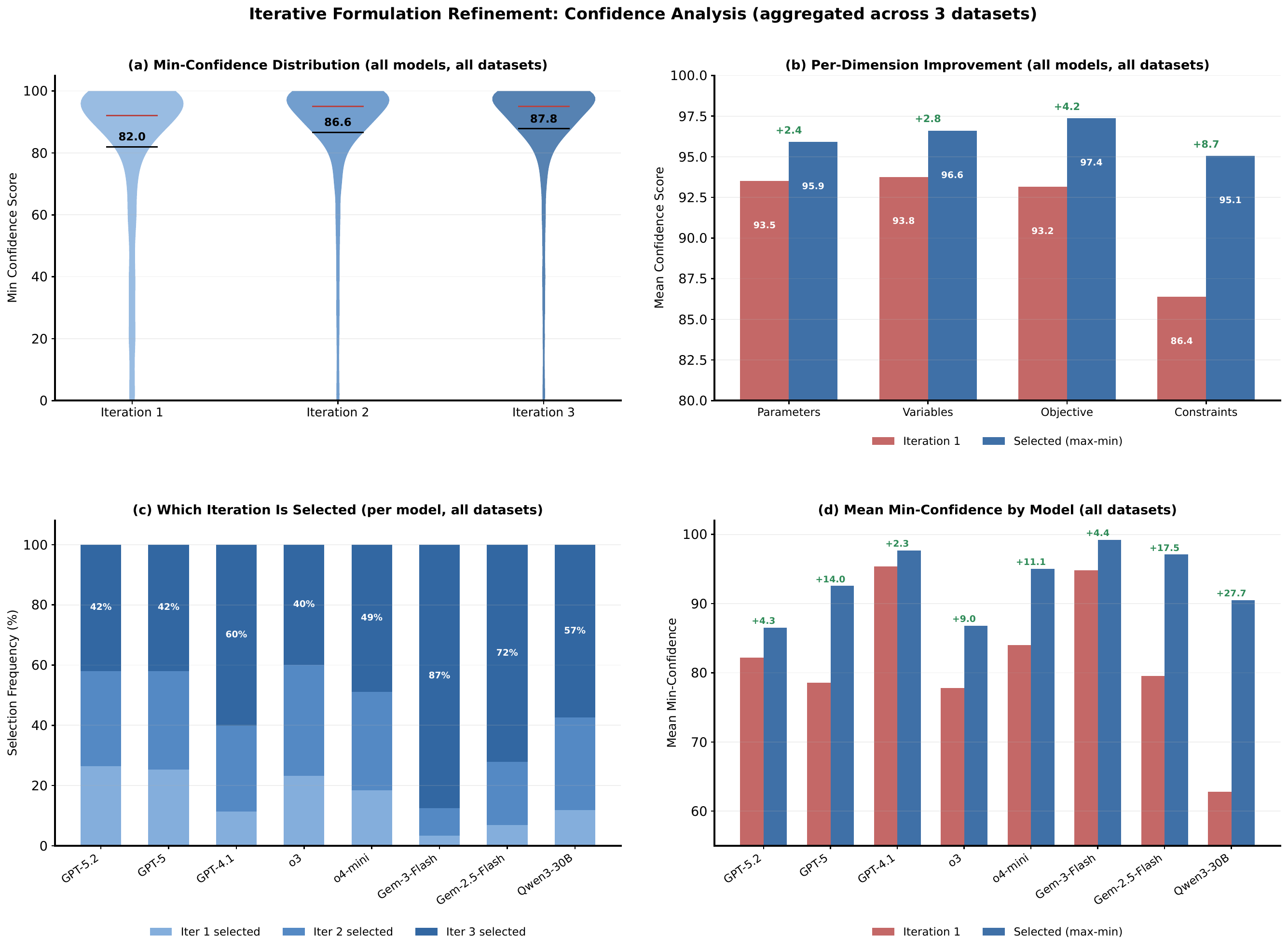}
  \caption{Confidence analysis of iterative refinement (aggregated across 3 datasets). (a)~Min-confidence increases across iterations (82.0 $\to$ 87.8). (b)~Per-dimension improvement from Iteration~1 to the selected candidate: constraints improve the most (+8.7), followed by objective (+4.2), variables (+2.8), and parameters (+2.4). (c)~Selection frequency shifts toward later iterations on weaker backbones. (d)~Selected candidates consistently have higher min-confidence than the first candidate, with the gap largest on weaker backbones.}
  \label{fig:confidence_4panel}
\end{figure*}

Figure~\ref{fig:confidence_4panel} summarizes how confidence changes across iterations. Min-confidence rises from 82.0 to 87.8 (panel a). From the first candidate to the selected one, constraints improve the most (+8.7), followed by objective (+4.2), variables (+2.8), and parameters (+2.4) (panel b), matching the error profile in Section~\ref{sec:modeling_errors}. Later iterations are selected more often on weaker models (panel c), and the selected candidate's min-confidence is consistently higher than the first candidate's (panel d), with the largest gaps on Qwen3-30B (+27.7) and Gemini-2.5-Flash (+17.5).

\textbf{Experiment 1: Max-min vs.\ first-candidate selection.} Table~\ref{tab:ablation_results} already gives this comparison: solving the max-min selected candidate instead of the first candidate improves accuracy on 7 of 8 backbones, for a +3.0-point cross-model mean gain. This indicates that the confidence criterion adds value beyond candidate diversity.

\begin{figure}[t]
  \centering
  \includegraphics[width=0.8\linewidth]{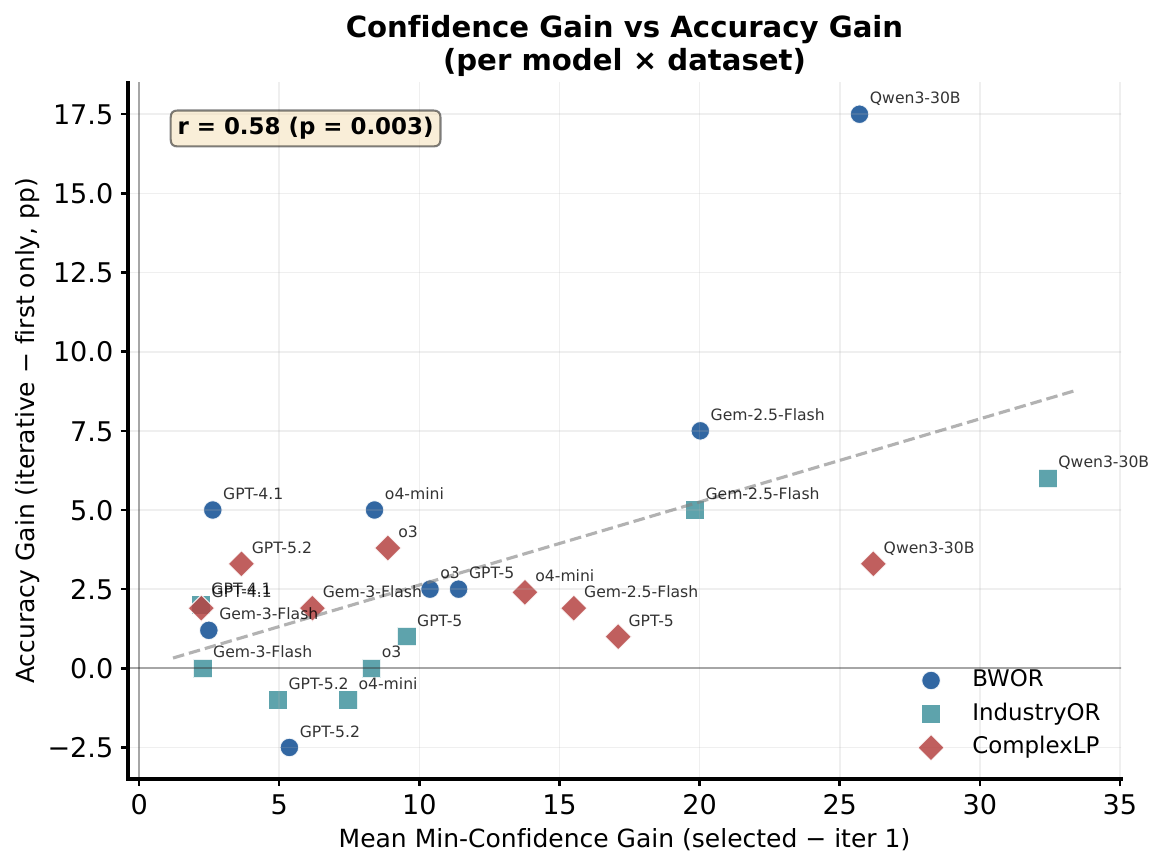}
  \caption{Confidence gain vs.\ accuracy gain per model--benchmark pair. Each point represents one (model, dataset) combination. The positive correlation ($r = 0.58$, $p = 0.003$) confirms that confidence improvements from iterative modeling translate into accuracy improvements.}
  \label{fig:conf_vs_acc}
\end{figure}

\textbf{Experiment 2: Confidence gain predicts accuracy gain.} We plot the mean min-confidence gain (selected minus Iteration~1) against the accuracy gain (w/ iteration minus w/o iteration) for each model--benchmark pair (Figure~\ref{fig:conf_vs_acc}). The Pearson correlation is $r = 0.58$ ($p = 0.003$), indicating that larger confidence gains usually translate into larger accuracy gains.

BWOR shows the strongest payoff from confidence gains, while ComplexLP is noisier and includes several near-zero or negative gains despite higher confidence, consistent with its greater difficulty. Overall, the max-min criterion yields 181 beneficial overrides and 95 harmful ones across all problem--backbone pairs (1.9:1), with net-positive selection on all 8 backbones and all 3 benchmarks.

\subsection{Solver Dispatch Statistics} \label{sec:solver_stats}

\begin{table}[t]
  \centering
  \caption{Distribution of optimizer agent calls per model and dataset. Each cell shows Math / Comb / Meta /
  General counts.}
  \label{tab:optimizer-calls-v22}
  \resizebox{\textwidth}{!}{%
  \begin{tabular}{l|ccc|ccc|ccc|ccc}
  \toprule
  & \multicolumn{3}{c|}{\textbf{complexlp} (211)} & \multicolumn{3}{c|}{\textbf{industryor} (100)} &
  \multicolumn{3}{c|}{\textbf{BWOR} (80)} \\
  \textbf{Model} & Math & Comb & Meta/Gen & Math & Comb & Meta/Gen & Math & Comb & Meta/Gen \\
  \midrule
  GPT-5.2       & 199 & 12 & 0/0 & 95 & 4  & 0/1 & 78 & 2  & 0/0 \\
  GPT-5         & 207 & 4  & 0/0 & 95 & 4  & 1/0 & 76 & 4  & 0/0 \\
  GPT-4.1       & 178 & 33 & 0/0 & 90 & 9  & 1/0 & 67 & 13 & 0/0 \\
  o3            & 188 & 23 & 0/0 & 93 & 5  & 0/2 & 74 & 4  & 1/1 \\
  o4-mini       & 174 & 31 & 2/4 & 91 & 7  & 1/1 & 65 & 14 & 0/1 \\
  Gemini-3-Flash & 211 & 0  & 0/0 & 96 & 3  & 1/0 & 76 & 4  & 0/0 \\
  Gemini-2.5-Flash & 197 & 14 & 0/0 & 97 & 2  & 1/0 & 76 & 4  & 0/0 \\
  Qwen3-30B     & 175 & 33 & 0/3 & 87 & 10 & 0/3 & 64 & 13 & 1/2 \\
  \bottomrule
  \end{tabular}%
  }
\end{table}

Table~\ref{tab:optimizer-calls-v22} shows that the mathematical optimizer dominates the workload: across 3{,}128 model--problem invocations, it handles 91.1\% of calls, versus 8.0\% for the combinatorial optimizer and less than 1\% combined for the metaheuristic and general optimizers. This largely reflects the benchmarks, which are overwhelmingly LP/IP/MIP. Routing still varies by backbone: GPT-4.1, o4-mini, and Qwen3-30B send 13--15\% of problems to the combinatorial optimizer, while Gemini-3-Flash sends fewer than 2\%.

The low use of non-mathematical optimizers limits empirical validation of solver dispatch. With only 253 combinatorial, 10 metaheuristic, and 18 general-optimizer calls, we cannot reliably test whether specialized routing improves accuracy over always using a mathematical programming solver. The contribution here is therefore architectural rather than strongly empirical, and broader validation will require more diverse benchmarks.

\subsection{Cost and Efficiency} \label{sec:cost}

COOPA's multi-step pipeline is inherently more expensive than single-pass methods such as OR-LLM-Agent. To measure that overhead, we sample 10 problems from each benchmark (30 total), run all five methods with Gemini-2.5-Flash, and record wall-clock time, total tokens, API calls, and estimated API cost. Table~\ref{tab:cost} reports the averages.

\begin{table}[h]
\centering
\caption{Cost and efficiency metrics averaged over 30 randomly sampled problems (10 per benchmark) using Gemini-2.5-Flash. Wall-clock time includes all LLM calls and code execution.}
\label{tab:cost}
\begin{tabular}{lcccc}
\toprule
\textbf{Method} & \textbf{Wall Time (s)} & \textbf{Total Tokens (K)} & \textbf{API Calls} & \textbf{Cost (\$)} \\
\midrule
Chain-of-Experts & 101.7 & 26.0 & 7.0 & 0.044 \\
OptiMUS & 198.8 & 88.7 & 36.4 & 0.103 \\
OptiTree & 74.0 & 15.6 & 3.5 & 0.028 \\
OR-LLM-Agent & 61.3 & 9.8 & 2.1 & 0.020 \\
\textbf{COOPA (Ours)} & 201.4 & 146.0 & 14.5 & 0.138 \\
\bottomrule
\end{tabular}
\end{table}

COOPA is the most expensive method on all four metrics: \$0.138 per problem on average, about $7\times$ OR-LLM-Agent (\$0.020) and $3\times$ CoE (\$0.044). The main driver is token usage: 146K tokens per problem, reflecting $k=3$ candidate generations and confidence evaluations. COOPA uses 14.5 API calls per problem, fewer than OptiMUS (36.4) but far more than OR-LLM-Agent (2.1) and OptiTree (3.5). Its wall-clock time (201.4s) is similar to OptiMUS (198.8s) and about $3\times$ OR-LLM-Agent (61.3s).

In absolute terms, the cost remains modest: a 100-problem benchmark costs under \$14 with COOPA. That premium may be acceptable when a wrong formulation is expensive, but it is less attractive in high-throughput settings. In those settings, the ablation in Section~\ref{sec:ablation} suggests that disabling iterative modeling can lower cost while retaining structured output and solver dispatch.

Wall-clock time is sensitive to API latency, so token counts and API calls are the more reliable cross-method comparisons.


\section{Discussion and Limitations} \label{sec:discussion}

\subsection{Modularity} \label{sec:disc_modularity}

A central design principle of COOPA is that each workflow component is independent and replaceable. This modularity has direct practical value: adding support for a new solver or problem class requires only defining a new optimizer agent (a system prompt and tool access list), with no changes to the other architectural components. For example, integrating a constraint programming solver (e.g., MiniZinc) or a stochastic programming framework (e.g., PySP) would require a single new agent definition. Fine-tuned models from the training-based literature (e.g., ORLM \cite{tang2024orlm}, SIRL \cite{chen2025solver}) could similarly be integrated as specialized backbone LLMs within individual agents, combining the benefits of training-based and pipeline approaches.

The scalable multi-solver architecture addresses a real gap in existing systems. As discussed in Section~\ref{sec:solver_dispatch}, problems such as multi-objective portfolio optimization, where the goal is to compute a Pareto front of non-dominated solutions balancing return and risk, are often more naturally handled by multi-objective evolutionary algorithms (e.g., NSGA-II via pymoo) than by repeated scalarization with a single mathematical-programming solver. COOPA's architecture supports such problems through its metaheuristic optimizer agent without forcing them into a single LP/MILP-style formulation. Appendix~\ref{sec:appendix_car_side_impact} demonstrates this on a real engineering design problem.

We acknowledge that the multi-solver dispatch is the least empirically validated design choice in this paper. The solver dispatch statistics (Table~\ref{tab:optimizer-calls-v22}) show that 91.1\% of problems are routed to the Mathematical Optimizer, reflecting the LP/MILP focus of existing benchmarks rather than the architectural capability. The qualitative case study in Appendix~\ref{sec:appendix_car_side_impact} provides one demonstration, but a comprehensive evaluation would require benchmarks with greater problem-type diversity, including vehicle routing, job-shop scheduling, and multi-objective optimization at meaningful scale.

\subsection{Interpretability} \label{sec:disc_interpretability}

COOPA provides two layers of interpretability that existing methods lack: (1)~source traceability linking each modeling element to quoted problem text, and (2)~confidence scores with natural-language explanations articulating the LLM's uncertainty. These features are practically important because OR practitioners need to trust and verify automated models before deploying solutions in high-stakes settings; a supply chain planner, for instance, must confirm that constraints correctly capture capacity limits and demand forecasts.

We note that the interpretability contribution in this paper is primarily architectural: we demonstrate that the system produces source references and confidence explanations, and we illustrate their use in the case studies (Section~\ref{sec:case_studies}). We do not empirically measure whether these features improve human verification efficiency or error detection rates, nor do we separately score the factual accuracy and completeness of the extracted source references. Controlled user studies and source-level audits with OR practitioners would substantially strengthen this contribution and are an important direction for future work.

\subsection{Ablation Takeaways} \label{sec:disc_ablation}

The ablation results (Table~\ref{tab:ablation_results}) reveal that iterative confidence-based modeling is the primary driver of COOPA's advantage over baselines. Without iterative modeling, COOPA's cross-model mean is 61.8\%, comparable to OR-LLM-Agent (61.6\%) and OptiTree (61.3\%). Iterative modeling adds 3.0 percentage points on average (61.8\% $\to$ 64.8\%), accounting for nearly all of COOPA's lead over the next-best baseline. This finding confirms the hypothesis from Section~\ref{sec:modeling_errors}: formulation quality is the central bottleneck in LLM-based OR, and a mechanism that enables the LLM to evaluate and refine its own formulations addresses this bottleneck directly.

The ablation deltas are inversely correlated with backbone capability. Qwen3-30B, the weakest model, gains the most (+8.9pp), while GPT-5.2, the strongest, is essentially unchanged ($-$0.1pp). This pattern suggests that iterative modeling functions as an equalizer: it compensates for the lower single-attempt quality of weaker backbones by providing multiple chances to identify and correct formulation errors. For stronger models that already produce high-quality first candidates, the mechanism still provides modest gains (+1.1 to +2.1pp on GPT-5, o3, Gemini-3-Flash) rather than degradation, indicating that the confidence evaluator generally avoids overriding correct formulations even when the margin for improvement is small.

For practitioners building LLM-based OR tools, the ablation supports deploying iterative modeling broadly across backbone models. The consistent gains on 7 of 8 backbones indicate that the mechanism is robust rather than narrowly tuned to specific model families. The primary consideration is cost rather than accuracy risk: iterative modeling increases per-problem cost by approximately $7\times$ (Section~\ref{sec:cost}), so practitioners with tight cost budgets may prefer the single-candidate variant, which already matches the best existing baselines. When accuracy is the priority, iterative modeling is justified across all tested backbones.

We emphasize that this is a \emph{within-system} ablation: it measures the contribution of iterative modeling within COOPA's architecture, not across different systems. The gains from iterative modeling depend on the structured output format that enables meaningful confidence evaluation; whether similar gains would arise in architectures with different output representations remains an open question.

\subsection{Confidence Reliability} \label{sec:disc_confidence}

The max-min selection criterion is only as effective as the LLM's ability to accurately self-evaluate its formulations. Two failure modes are possible: systematic overconfidence (high scores for incorrect formulations) and poor discrimination (similar scores for correct and incorrect formulations).

The ablation results provide indirect evidence that the confidence-based selection criterion is functional: iterative modeling improves accuracy on 7 of 8 backbones, with gains of up to +8.9 percentage points. If confidence scores were uninformative, generating additional candidates would provide only diversity-based improvement (roughly proportional to the chance that any one of $k$ candidates is correct), and the max-min criterion would perform no better than random selection. The consistently positive deltas suggest that the evaluator provides genuine signal for distinguishing better formulations from worse ones. However, the magnitude of the gains varies substantially across backbones, from +8.9pp on Qwen3-30B to $-$0.1pp on GPT-5.2. This variation raises the question of whether the evaluator is well-calibrated in absolute terms, or whether it merely provides sufficient relative ranking to support selection. The targeted analyses in Appendix~\ref{sec:calibration} address this question directly.

To move beyond indirect evidence, we conduct three targeted analyses in Appendix~\ref{sec:calibration}. First, we compare the accuracy of max-min selection against always using the first candidate, testing whether the criterion adds value beyond candidate diversity. Second, we measure whether confidence gains predict accuracy gains across model--benchmark pairs. Third, we examine cases where the criterion overrides the first candidate, measuring how often such overrides are beneficial versus harmful. Together, these analyses determine whether the confidence-based selection mechanism provides genuine signal or operates as a near-random selector among candidates.

\subsection{Potential Impacts} \label{sec:disc_impacts}

COOPA could lower the expertise barrier for optimization modeling in settings such as logistics, energy, manufacturing, and public-service planning, which may improve access to OR tools beyond specialist teams. At the same time, incorrect automatically generated formulations could misallocate resources, hide important tradeoffs, or create unjustified confidence in flawed decisions if used without human review.

For this reason, we view COOPA as a decision-support system rather than an autonomous decision-maker. Its source traceability, confidence explanations, and explicit case studies are intended to support human checking before deployment, especially in high-stakes settings where modeling errors can have operational or social consequences.

\end{document}